\PassOptionsToPackage{table}{xcolor}
\PassOptionsToPackage{xcdraw}{xcolor}

\documentclass[a4paper,10pt]{article}

\usepackage{amsthm}
\RequirePackage{thmtools}

\declaretheorem[name=Theorem]{theorem}

\declaretheorem[name=Proposition, numberlike=theorem]{proposition}
\declaretheorem[name=Corollary, numberlike=theorem]{corollary}
\declaretheorem[name=Definition]{definition}
\declaretheorem[name=Example]{example}

\usepackage{fullpage}

\usepackage{microtype}
\usepackage{graphicx}
\usepackage{subcaption}
\usepackage{booktabs}
\usepackage{hyperref}
\usepackage{authblk}

\usepackage[numbers, sort&compress]{natbib}

\usepackage{url}
\usepackage{researchpack}
\usepackage[capitalize,nameinlink]{cleveref}
\usepackage{multirow}
\usepackage{enumitem}
\usepackage{tablefootnote}
\usepackage{comment}
\usepackage{graphicx}
\usepackage[makeroom]{cancel}
\usepackage{subcaption}
\usepackage{booktabs}
\usepackage{algorithm}
\usepackage[noend]{algpseudocode}
\usepackage{wrapfig}
\usepackage{tcolorbox}

\hypersetup{
    colorlinks=true,
    linkcolor=blue,
    citecolor=blue,
    urlcolor=blue
}

\definecolor{VibrantGreen}{RGB}{0, 204, 0}

\usepackage{pifont}

\newcommand{\vPa}{\ensuremath{\mathbf{Pa}}\xspace}

\newcommand{\Caused}{\ensuremath{\coloneqq}\xspace}

\newcommand{\dprod}[2]{\ensuremath{\langle #1, #2 \rangle}\xspace}
\newcommand{\expect}[1]{\ensuremath{\bbE \left[ #1 \right]}\xspace}
\newcommand{\abs}[1]{\ensuremath{\left| #1 \right|}\xspace}

\newcommand{\SAR}{{\tt SAR}\xspace}
\newcommand{\CAR}{{\tt CAR}\xspace}

\newcommand{\TSAR}{{\tt T-SAR}\xspace}
\newcommand{\TCAR}{{\tt T-CAR}\xspace}
\newcommand{\IMF}{{\tt IMF}\xspace}

\newcommand{\GDT}[1]{\textcolor{black}{#1}\xspace}

\newcommand{\evidenzia}[1]{\textit{#1}\xspace}

\newcommand{\Yes}{\raisebox{-0.2em}{\includegraphics[width=2.5ex]{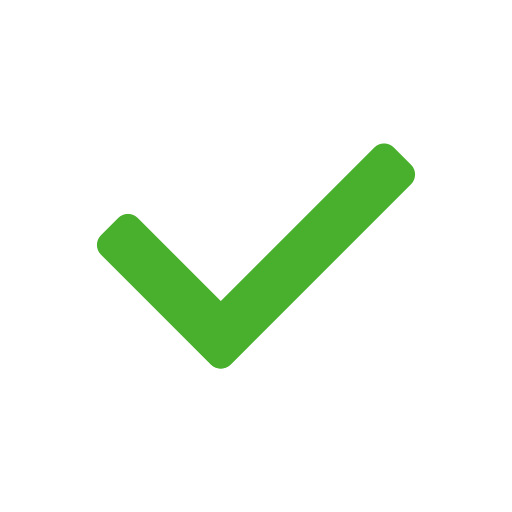}}\xspace}
\newcommand{\No}{\raisebox{-0.2em}{\includegraphics[width=2.5ex]{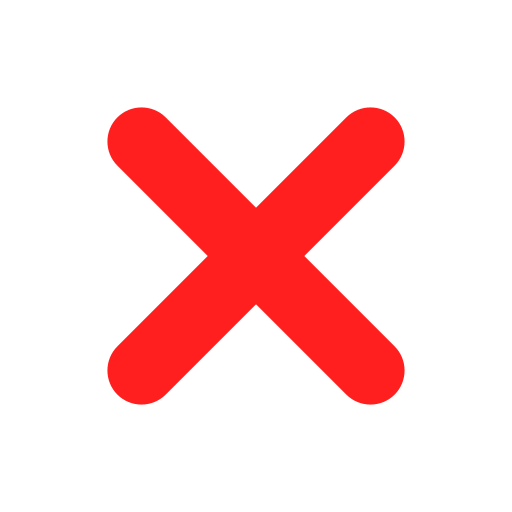}}\xspace}
\newcommand{\Hammer}{\raisebox{-0.2em}{\includegraphics[width=2.5ex]{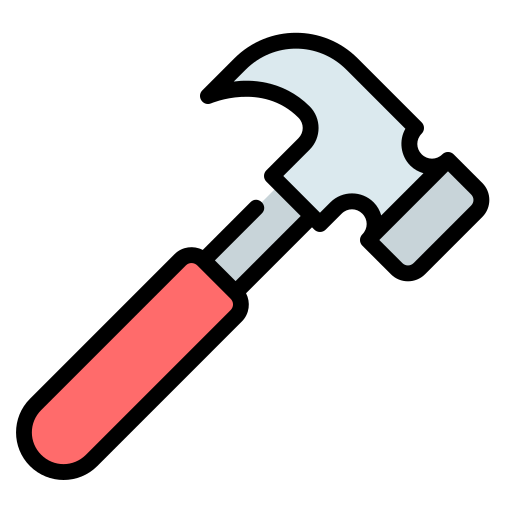}}\xspace}
\newcommand{\Users}{\raisebox{-0.2em}{\includegraphics[width=7.5ex]{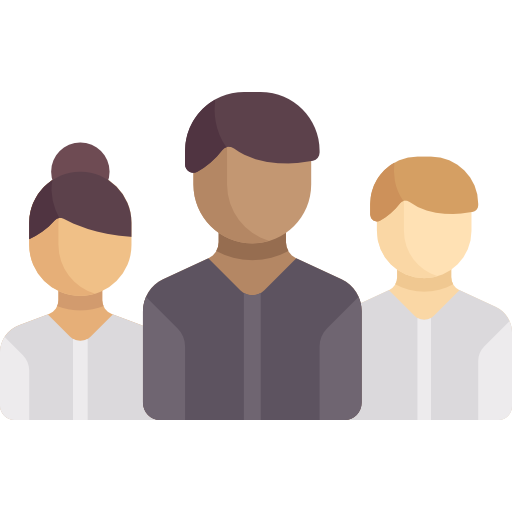}}\xspace}

\usepackage{tikz}
\usetikzlibrary{shapes,shapes.callouts,decorations,decorations.pathmorphing,shapes,arrows,calc,arrows.meta,fit,positioning,backgrounds}
\tikzset{
    -Latex,auto,node distance =0.5 cm and 0.5 cm,semithick,
    state/.style ={circle, draw, minimum size = 1.0 cm,inner sep=0.7pt, font=\Large},
    point/.style = {circle, draw, inner sep=0.04cm,fill,node contents={}},
    bidirected/.style={Latex-Latex,dashed},
    el/.style = {inner sep=2pt, align=left, sloped}
}

\title{Time Can Invalidate Algorithmic Recourse}

\author[1]{Giovanni De Toni} 
\author[2,3]{Stefano Teso} 
\author[1]{Bruno Lepri} 
\author[2]{Andrea Passerini} 

\affil[1]{Fondazione Bruno Kessler, \texttt{\{gdetoni,lepri\}@fbk.eu}}
\affil[2]{DISI, University of Trento, \texttt{andrea.passerini@unitn.it}}
\affil[3]{CIMeC, University of Trento, \texttt{stefano.teso@unitn.it}}

\date{\vspace{-10mm}}

\begin{document}

\maketitle

\begin{abstract}
Algorithmic Recourse (AR) aims to provide users with actionable steps to overturn unfavourable decisions made by machine learning predictors.
However, these actions often take time to implement (\eg getting a degree can take years), and their effects may vary as the world evolves.
Thus, it is natural to ask for recourse that remains valid in a dynamic environment.
In this paper, we study the robustness of algorithmic recourse over time by casting the problem through the lens of causality.
We demonstrate theoretically and empirically that (even robust) causal AR methods can fail over time except in the -- unlikely -- case that the world is \textit{stationary}.
Even more critically, unless the world is fully deterministic, \textit{counterfactual} AR cannot be solved optimally.
To account for this, we propose a simple yet effective algorithm for temporal AR that explicitly accounts for time under the assumption of having access to an estimator approximating the stochastic process.
Our simulations on synthetic and realistic datasets show how considering time produces more resilient solutions to potential trends in the data distribution.
\end{abstract}

\section{Introduction}

Machine Learning (ML) models play an increasingly prominent role in high-stakes decision-making tasks such as credit lending \cite{barbaglia2021forecastingloan}, bail approval \cite{dressel2018accuracy}, and medical diagnosis \cite{yoo2019adopting}.
\GDT{
Over the past years, concerns have risen regarding the safety and fairness of these systems \citep{grgichlaca2018human, binns2018justicepercetion}, as well as their impact on human dignity and autonomy \citep{jobin2019global, kaminski2018binary}.
Indeed, there have been examples of algorithmic systems providing discriminatory assessments in contexts such as crime justice \citep{angwin2022machine}, employment decisions \citep{turque2012creative}, and benefit allocations \citep{heikkila2021dutchscandal}.
}
Consequently, there is a growing consensus that algorithmic systems must offer users the means to challenge their decisions, thus preserving human agency - a principle now reflected in recent AI legislation such as the GDPR and the European AI Act \cite{com2021laying,council2024regulation}.
\evidenzia{Algorithmic Recourse} (AR) \cite{wachter2017counterfactual} aims to identify counterfactual explanations that users can follow to overturn unfavourable machine decisions.
\GDT{A counterfactual explanation highlights \textit{``the smallest change to the feature values that changes the prediction to a predefined output''} \cite{guidotti2024counterfactual}.}
\GDT{For example, in a credit lending task}, a counterfactual recourse might suggest a user obtain a \GDT{Master's degree} as this will net them a higher income, thereby improving loan approval prospects.

\begin{figure*}[t]
    \centering
    \begin{minipage}[c]{0.35\linewidth} 
        \centering
        \includegraphics[width=\linewidth]{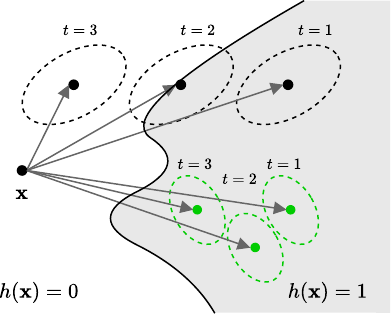}
        \subcaption{Recourse validity evolves over time}
        \label{fig:setting}
    \end{minipage}
    \hfill
    \begin{minipage}[c]{0.6\linewidth} 
        \centering
        \resizebox{\textwidth}{!}{%
\begin{tikzpicture}[node distance=2cm, >=stealth, auto]

    \node[state, line width=0.5mm, fill=white] (X1_prev) at (0, 1.5) {$A^{t}$};
    \node[state, line width=0.5mm, fill=black!10] (X2_prev) at (-1, 0) {$I^{t}$};
    \node[state, line width=0.5mm, fill=black!10] (X3_prev) at (1, 0) {$D^{t}$};

    \node[state, line width=0.5mm, fill=white] (X1_next) at (4, 1.5) {$A^{t+1}$};
    \node[state, line width=0.5mm, fill=black!10] (X2_next) at (3, 0) {$I^{t+1}$};
    \node[state, line width=0.5mm, fill=black!10] (X3_next) at (5, 0) {$D^{t+1}$};

    \node[state, line width=0.5mm, fill=white] (X1_next2) at (8, 1.5) {$A^{t+2}$};
    \node[state, line width=0.5mm, fill=black!10, draw=red] (X2_next2) at (7, 0) {$I^{t+2}$};
    \node[state, line width=0.5mm, fill=black!10] (X3_next2) at (9, 0) {$D^{t+2}$};

    \node[shape=rectangle, text width=5cm, align=center, line width=0.1mm] (Hammer) at (7, 3) {\Large \Hammer $do(I^{t+2} = i^{t+2} + \theta)$};

    \begin{pgfonlayer}{background}
    \shade[left color=blue!0, right color=blue!30]  
        ($(X1_next.north west)+(-2,0.5)$) rectangle 
        ($(X3_next2.south east)+(0.7,-0.8)$);

    \node[anchor=south west, font=\large, text=black] (USERS)
        at ($(X1_prev.north west)+(-2,-0.2)$) {\Users};
    
    \node[draw, fill=white!30, font=\normalsize, rectangle callout, callout absolute pointer={(USERS)+ (0.3,0.3)}, align=center] 
        at (1, 3.7) {Our loan application was denied at time $t$. \\ \textit{How can we obtain it at time $t+2$?}};
    
    \end{pgfonlayer}

    \draw[>=Stealth, line width=0.3mm] (X1_prev) -- (X2_prev);
    \draw[>=Stealth, line width=0.3mm] (X2_prev) to (X3_prev);
    \draw[>=Stealth, line width=0.3mm] (X1_prev) -- (X3_prev);

    \draw[>=Stealth, line width=0.3mm] (X1_next) -- (X2_next);
    \draw[>=Stealth, line width=0.3mm] (X2_next) -- (X3_next);
    \draw[>=Stealth, line width=0.3mm] (X1_next) -- (X3_next);

    \draw[>=Stealth, line width=0.3mm] (X1_next2) -- (X2_next2);
    \draw[>=Stealth, line width=0.3mm] (X2_next2) -- (X3_next2);
    \draw[>=Stealth, line width=0.3mm] (X1_next2) -- (X3_next2);

    \draw[>=Stealth,  line width=0.3mm] (X1_prev) -- (X1_next);
    \draw[>=Stealth,  line width=0.3mm] (X1_next) -- (X1_next2);

    \draw[>=Stealth,  line width=0.3mm] (X2_prev) to [bend right] (X2_next);
    \draw[>=Stealth,  line width=0.3mm] (X2_next) to [bend right] (X2_next2);

    \draw[>=Stealth,  line width=0.3mm] (X3_prev) to [bend right] (X3_next);
    \draw[>=Stealth,  line width=0.3mm] (X3_next) to [bend right] (X3_next2);

    \draw[>=Stealth,  line width=0.3mm] (-2, 1.5) -- (X1_prev);
    \draw[>=Stealth,  line width=0.3mm] (-2, 0) -- (X2_prev);
    \draw[>=Stealth,  line width=0.3mm] (-2, -0.5) to [bend right] (X3_prev);

    \draw[>=Stealth,  line width=0.3mm] (X1_next2) -- (10.0, 1.5);
    \draw[>=Stealth,  line width=0.3mm] (X2_next2) to [bend right] (10.0, -0.5);
    \draw[>=Stealth,  line width=0.3mm] (X3_next2) -- (10.0, 0);

    \draw[>=Stealth, decorate, line width=0.3mm, color=red, decoration={snake, amplitude=0.5mm}] (Hammer) -- (X2_next2);

\end{tikzpicture}
}
        \subcaption{Causal time series with actionable and non-actionable features}
        \label{fig:temporal-causal-graph}
    \end{minipage} 
    \caption{(a) A recourse suggestion may fail to achieve a positive outcome ($h(\vx) = 1$) over time, even when accounting for the uncertainty surrounding the individual $\vx$ requesting recourse (\textbf{black}). In contrast, a temporally robust recourse suggestion ensures a positive classification, regardless of the time $t$ at which it is applied (\textbf{\textcolor{VibrantGreen}{green}}) (b) Illustration of a time series described by a fictional causal graph. Each node represents the value of the $j$-th feature at time $t$, categorized as either actionable ( \raisebox{0.5ex}{\fcolorbox[HTML]{000000}{BDBDBD}{\rule{0pt}{1pt}\rule{1pt}{0pt}}} $I$: \textit{income}, $D$: \textit{debt-to-income ratio}) or non-actionable ( \raisebox{0.5ex}{\fcolorbox[HTML]{000000}{FFFFFF}{\rule{0pt}{1pt}\rule{1pt}{0pt}}} $A$: \textit{age}). Arrows depict temporal causal relationships between variables, while nodes showing future timesteps are shown against a progressively bluer background. 
    Similar graphs are used in our simulations (\cf \cref{sec:experiments}).
    In temporal algorithmic recourse (\cref{def:tar-definition}), user actions are modelled as \textit{robust interventions} (\Hammer), modifying one or more actionable features (\textbf{\textcolor{red}{red}}) to ensure recourse at a specific future $t+\tau$.}
    \label{fig:combined}
\end{figure*}

In order to be of use, suggested recourse must be \textit{actionable} \cite{ustun2019actionable} and sufficiently inexpensive for the user to implement \cite{detoni2024personalized}.
We argue that actionability inherently includes considerations of \textit{timing}.
Specifically:
(\textit{i}) implementing and realizing the effects of recourse actions takes time, and
(\textit{ii}) the same action performed at different times may yield different outcomes.
\GDT{For instance, earning a Master's degree requires several years and only begins to influence salary after some delay. Additionally, pursuing a degree later in life may impose a substantial financial burden while providing a comparatively smaller salary increase, often due to a \textit{ceiling effect} \cite{lattimore2023economic}.}
\GDT{In realistic mortgage applications data, real features that a counterfactual might suggest modifying include the debt-to-income ratio, acquiring collateral, or addressing insufficient liquidity \cite{bhutta2017residential}. Yet again, improving such features demands significant effort and time.%
\footnote{\GDT{While this paper presents examples from the financial sector, similar challenges arise in other domains, such as the labour market \cite{greenEuropesEvolvingGraduate2021} and college admissions \cite{hossler2019study}, all of which are increasingly influenced by AI-driven decision systems. These domains also have a potential need for algorithmic recourse recommendations, which must account for trends (\eg evolving job market skills) and the timing of users’ ability to act on these recommendations.}}
}
Our key insight is that, since \evidenzia{time plays a key role in the effectiveness of recourse}, one has to ensure that recourse suggestions should be \textit{\GDT{robust to temporal factors}}.
\GDT{Specifically, recourse should lead to a favourable outcome regardless of when the user acts or, at a minimum, at a user-defined future point. \Cref{fig:setting} illustrates this concept.}

\GDT{While many studies focus on improving the robustness of recourse in various contexts \cite{jiang2024surveyrobustness}, to the best of our knowledge, little has been done to formalize and study recourse robustness over time explicitly.}
For instance, \citet{upadhyay2021towards} examines recourse under model shifts due to retraining, \citet{pawelczyk2022probabilistically} addresses scenarios where users imperfectly execute recourse actions, and \citet{dominguez2022adversarial} explores robustness against input specification errors. However, explicit consideration of time as a factor in recourse robustness remains limited.
\citet{beretta2023importance} recently incorporated time into the recourse cost but did not address its impact on the validity of recourse recommendations.

In this paper, we study the problem of time in algorithmic recourse through the lens of \textit{causality} \cite{pearl2009causality}.
In \textit{causal algorithmic recourse} \cite{karimi2020algorithmic, dominguez2022adversarial, majumdar2024carma, beretta2023importance}, recourse suggestions are modelled as \textit{interventions} on the user's features, thus giving a reliable representation of how the features will \GDT{hypothetically} change as the user acts on them to achieve recourse, provided we know the (approximate) \textit{causal model} \cite{karimi2020algorithmic}.
We consider a novel setting where we are asked to provide recourse for a causal discrete-time stochastic process characterized by trends. \GDT{An example of this setting is depicted in \Cref{fig:temporal-causal-graph}.}
Our \GDT{theoretical and empirical results} challenge the usefulness of the mainstream variants of causal and non-causal AR when extended over time, showing their recommendations \evidenzia{can become invalid} as time progresses.

\textbf{Our contributions.} Summarizing, we (i) introduce a sound but intuitive formalization of temporal causal AR (\cref{def:tar-definition}), based on causality \cite{pearl2009causality} and time-series with independent noise \cite{peters2013causal},
(ii) show theoretically how uncertainty and non-stationarity hinder optimal counterfactual and sub-population recourse (\CAR and \SAR, \cite{karimi2020algorithmic}) for simple discrete-time stochastic processes,
(iii) show how robustifying causal recourse via \textit{uncertainty sets}~\cite{dominguez2022adversarial} is not enough to counteract time (\cref{prop:robust-recourse-fails-over-time}),
(iv) present numerical simulations showcasing the detrimental effects of time on robust (non-)causal AR approaches, and (v) \GDT{propose a simple time-aware algorithm (\Cref{alg:differentiable-temporal-recourse}) that mitigates these issues in synthetic and realistic scenarios (\Cref{sec:experiments}).}

\section{Preliminaries and Related Work}
\label{sec:preliminaries}

Throughout, we indicate (random) variables $X$ in upper case, constants $x$ in lower case, vectors in bold $\vx$, and sets $\calX$ in italics.  We also abbreviate $\{1, \ldots, n\}$ as $[n]$.

\textbf{Causality}. \textit{Structural Causal Models} (SCMs) \cite{pearl2009causality} allow us to formalize and reason about the causal behaviour of a system.
An SCM $\calM = (\vX, \vU, P, \calF)$ encompasses
endogenous variables $\vX = \{X_i\}^d_{i=1}$,
noise variables $\vU = \{U_i\}^d_{i=1}$ distributed according to $P(\vU)$,
and structural assignments $\calF$ of the form $X_i \Caused f_i(\vPa_i, U_i)$ that describe all causal relationships between variables and their \textit{direct causes} (or parents) $\vPa_i \subseteq \vX \setminus X_i$ \GDT{\eg the link between education level and income.}
An SCM induces a pushforward distribution $P(\vX, \vU) = P(\vX \mid \vU) P(\vU)$, where $P(\vX \mid \vU)$ is deterministic.
\textit{Hard interventions} $do(\vX_\calI=\vtheta)$ allow implementing external actions on an SCM \GDT{to simulate their causal effect on the system, \eg understanding the impact of getting either a PhD or a Master's on our salary.}
They replace a subset of variables $\vX_\calI \subseteq \vX$ with constants $\vtheta \in \bbR^{|\vX_\calI|}$, detached from their original parents, yielding a new SCM $\calM^{do(\vtheta)}$ with updated structural assignments $\calF^{do(\vtheta)}$ and an associated \evidenzia{interventional distribution} $P^{do(\vtheta)}(\vX)$.
Instead, \textit{soft interventions} $do(\vX_\calI=\vx_\calI + \vtheta)$ change how the affected variables depend on their parents without detaching them.
We shorten both kinds of intervention as $do(\vtheta)$ for readability.
%
SCMs also enable us to reason \textit{counterfactually} on what would have happened if the world were different due to an intervention $do(\vtheta)$, all else being equal. 
\GDT{For example, we could simulate questions such as \textit{``What would have happened if we had undertaken a PhD rather than a Master?''}.}
Given a realization $\vx$, the \evidenzia{counterfactual distribution} $P^{do(\vtheta), \vX=\vx}(\vX)$ is obtained by first \textit{abducing} the exogenous factors $\vU$ in the original SCM and then inferring the state of $\vX$ in the intervened SCM, that is, $P^{do(\vtheta), \vX=\vx}(\vX) = P^{do(\vtheta)}(\vX \mid \vU) P(\vU \mid \vX = \vx)$ \cite[Theorem~7.1.7]{pearl2009causality}.  If the structural equations are invertible, $P(\vU \mid \vx)$ is deterministic, and so is the counterfactual distribution.

\textbf{Causal algorithmic recourse}.  \GDT{In AR, the key quantities of interest are the \textit{user's state} $\vx \sim P(\vX)$, which encapsulates the user's features relevant to the decision-making task (\eg for a loan application: salary, debt-to-income ratio, occupation, etc.), and the \textit{outcome} $y \sim P(Y \mid \vX)$, representing the predicted event (\eg whether the user will repay their loan).}
The SCM underlying $P(\vX)$ is assumed to be known or estimated from data \cite{karimi2020algorithmic}, enabling us to apply interventions to evaluate the effect of changing the user's state while considering all causal dependencies between variables.
Given a (potentially black-box) classifier $h: \vx \mapsto [0,1]$ approximating $P(Y \mid \vX)$ and a realization $\vx$ yielding an undesirable outcome, \ie $h(\vx) < 1/2$, AR involves finding an intervention $\vtheta^*$ that, once implemented by the user, leads in expectation to a more favourable outcome.
There are two mainstream approaches to causal AR \cite{karimi2020algorithmic}.  \evidenzia{Sub-population recourse} (\SAR) provides recourse to users belonging to a specific sub-group, and it is defined as:\footnote{
The choice of $\vtheta$ is often restricted by actionability requirements (\eg age cannot be changed at will) or other constraints (\eg monotonicity: age can only increase).
We omit this detail for readability.
}
\begin{align}
    \textstyle
    \vtheta^* \in
    \argmin_{\vtheta \in \bbR^d} \ \bbE_{\hat{\vx}} [C(\hat{\vx}, \vx)]
    \quad
    \mathrm{s.t.}
    \quad
    \bbE_{\hat{\vx}}[h(\hat{\vx})] \geq 1/2
    \label{eqn:causal-recourse-equation}
\end{align}
Here, $C$ is a non-negative cost function measuring the user's effort, \eg the $\ell_2$-norm.
Selecting a user belonging to a specific subgroup amounts to sampling $\hat{\vx}$ in \cref{eqn:causal-recourse-equation} by conditioning on a subset of variables $P^{do(\vtheta)}(\vX \mid \vX_{nd(\calI)} = \vx_{nd(\calI)})$ where $nd(\calI)$ indicates the non-descendants of the intervened upon variables $\calI$.
Instead, \evidenzia{counterfactual recourse} (\CAR) allows computing \textit{individualized} recourse for the specific individual $\vx$.  It is formulated like \cref{eqn:causal-recourse-equation}, except that the interventional distribution $P^{do(\vtheta)}(\vX \mid \vX_{nd(\calI)} = \vx_{nd(\calI)})$ is replaced with the counterfactual distribution $P^{do(\vtheta), \vX=\vx}(\vX)$.
Lastly, since providing optimal recourse (\cref{eqn:causal-recourse-equation}) is harder for unrestricted SCMs \cite{karimi2020algorithmic}, causal AR typically assumes the SCM of $P(\vX)$ belongs to an identifiable and invertible class, \eg Additive Noise Models (ANMs) \cite{karimi2021algorithmic, dominguez2022adversarial, karimi2020model}.

\textbf{Further related works.}
\GDT{Our work extends the literature on causal algorithmic recourse \cite{majumdar2024carma, von2022fairness, karimi2020algorithmic, dominguez2022adversarial} and it builds upon research in robust algorithmic recourse, counterfactual explanations for time series, and causality.
The robustness literature focuses on generating counterfactual explanations that remain stable despite changes in the model $h$ \cite{ferrario2022robustness,pawelczyk2022trade,nguyendistributionally, meyer2023minimizing,upadhyay2021towards,hamman2023robust}, endogenous dynamics \cite{altmeyer2023endogenous}, model multiplicity \cite{pawelczyk20modelmulti, leofante2023counterfactual}, noisy intervention execution \cite{pawelczyk2022probabilistically,virgolin2023robustness}, and input instance uncertainty \cite{dominguez2022adversarial,slack2021cfcanmanipulated,artelt2021evaluating}.
We remark that none of these works explicitly formalize time as a dimension in their computational model.
Recently, \citet{fonseca2023setting} studied empirically algorithmic recourse in a multi-agent setting, focusing instead on the \textit{competition} for resources over time, without a causal notion.
The literature on counterfactual explanations for multivariate time series provides techniques to generate explanations for stochastic processes \cite{delaney2021instance, ates2021countime,yan2023self} by considering \textit{independently manipulable features} (IMF) (\ie no causal relationships between variables), Pearl causality \cite{pearl2009causality} or Granger causality \cite{granger1969investigating}. Nevertheless, practical causal inference over time remains an open research area, especially its conjunction with algorithmic recourse \cite{cinquini2024practical}.
Given the breadth and depth of the recourse and explainability fields, our discussion here is necessarily limited.
We refer interested readers to existing surveys \cite{jiang2024surveyrobustness,guidotti2024counterfactual,verma2020counterfactual,karimi2020survey} that cover many other aspects of these frameworks.
}

\begin{figure*}[t]
    \centering
    \includegraphics[width=\linewidth]{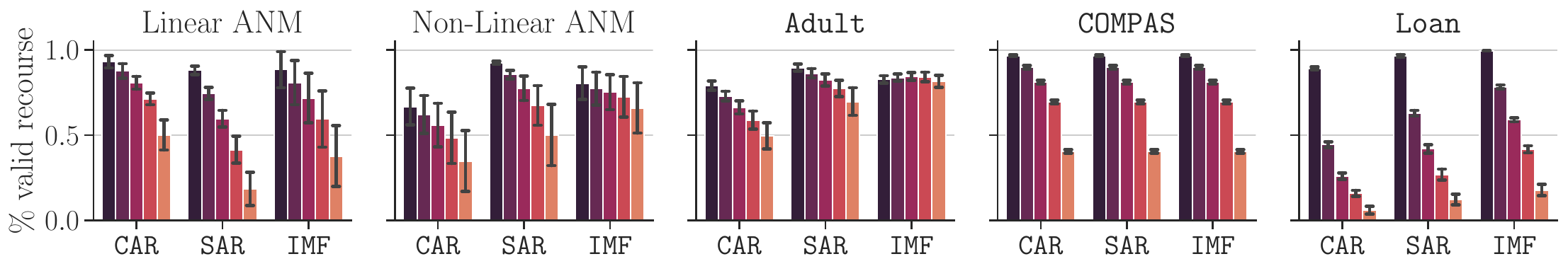}
    \caption{\textbf{Algorithmic recourse is not robust in time.} Empirical average validity and standard deviation of robust counterfactual (\CAR), sub-population (\SAR) and non-causal recourse (\IMF) at time $t=50$ for synthetic (\cref{sec:synthetic-experiment}) and realistic (\cref{sec:real-experiment}) time series with a non-linear trend. 
    We report the validity varying the strength $\alpha$ of the trend. Legend ($\alpha$):
    \raisebox{0.5ex}{\fcolorbox[HTML]{FFFFFF}{35193E}{\rule{0pt}{1pt}\rule{1pt}{0pt}}} $0$
    \raisebox{0.5ex}{\fcolorbox[HTML]{FFFFFF}{701F57}{\rule{0pt}{1pt}\rule{1pt}{0pt}}} $0.3$
    \raisebox{0.5ex}{\fcolorbox[HTML]{FFFFFF}{AD1759}{\rule{0pt}{1pt}\rule{1pt}{0pt}}} $0.5$
    \raisebox{0.5ex}{\fcolorbox[HTML]{FFFFFF}{E13342}{\rule{0pt}{1pt}\rule{1pt}{0pt}}} $0.7$
    \raisebox{0.5ex}{\fcolorbox[HTML]{FFFFFF}{F37651}{\rule{0pt}{1pt}\rule{1pt}{0pt}}} $1.0$.
    }
    \label{fig:non-robustness-causal-recourse}
\end{figure*}

\section{Algorithmic Recourse in Time}
\label{sec:temporal-algorithmic-recourse}%
In real-world applications, users do not implement nor complete suggested interventions immediately.  For instance, \GDT{improving our debt-to-income ratio to a satisfactory level can take months}. 
This is problematic because $P(\vX, Y)$ might change over time due to, \eg inflation rates, seasonality of loan interests, and classifier updates, meaning that \evidenzia{recourse produced by existing approaches could become ineffective or even counterproductive in the future}.
\cref{fig:non-robustness-causal-recourse} shows the empirical average validity (\% of interventions achieving recourse) of state-of-the-art robust (non-)causal recourse methods on different binary decision problems, where $P(\vX, Y)$ is a stochastic process exhibiting a non-linear trend.
Unfortunately, \evidenzia{current robust (non-)causal recourse methods are increasingly fragile over time proportionally to the trend's strength.}
\GDT{This section aims to provide a theoretical description and analysis of this issue. We present a summary of our findings in \cref{app:tab:methods-overview}, along with the characteristics of the (non-)causal recourse methods considered.}

\subsection{Formalizing Temporal Causal AR}
\label{sec:tcar}
Before proceeding, we need to extend causal AR with a time dimension.
We do so by considering a stochastic process $P(\vX^t, Y^t)$ capturing the evolution of the user's state and its relationship to the target variable over time $t \in \bbN$.
In the following, we assume that \textit{$P$ is induced by an SCM over the same variables} in which the parents of each variable lie in the past, and specifically within a fixed (but otherwise arbitrary) horizon $\rho \ge 1$, \ie $\vPa_{X_i^t} \subseteq \bigcup_{\delta=0}^\rho \vX^{t-\delta}$, and similarly for $Y^t$.
We also assume that the outcome $Y^t$ only depends on the user's states $\vX^1, \ldots, \vX^t$ up to time $t$.
We investigate the effect of time on recourse by considering stochastic processes where $P(\vX^t)$ might not be stationary, while the conditional distribution $P(Y^t \mid \vX^t)$ is unchanged, similarly to \textit{covariate shift} \cite{shimodaira2000improving}.
For this reason, we can assume the classifier $h^t$ is fixed.\footnote{
If this is not the case, one option is to leverage existing techniques for addressing changes due to retraining, such as \citet{upadhyay2021towards} and \citet{pawelczyk2022trade}.  These works and ours are complementary, and studying their interplay is a promising avenue for future work. 
}

\textbf{How do we describe time series with causality?} In the remainder we assume the stochastic process is a \textit{Time series Model with Independent Noise} (TiMINo), adapted from \cite{peters2013causal}:
\begin{definition}[TiMINo for Algorithmic Recourse]
    $P(\vX^t, Y^t)$ satisfies TiMINo if it causally factorizes as $X^t_i = f^t_{X_i}(\vPa_{X_i^t}) + U^t_{X_i}$ and $Y^t = f_Y^t(\vX^t) + U^t_{Y}$, for all $i \in [d]$, where $U_{X_i}^t, U_Y^t$ are jointly independent and identically distributed for all $i \in [d]$ and $t \in \bbN$.
    \label{eqn:scm-recourse-over-time-timino}
\end{definition}%
Under appropriate conditions, TiMINo SCMs are \textit{invertible}, allowing us to apply causal reasoning to infer the counterfactual distribution when computing AR, and can be identified from observational data.
Specifically, under appropriate choices of the family of $f^t_{X_i}$, $f^t_{Y}$ -- which still allow them to be non-linear -- and $P(\vU^t)$ we are guaranteed to identify both the \textit{summary graph} and \textit{full-time} graph \cite{peters2013causal}.
Moreover, \citet{peters2013causal} provides a causal discovery procedure for \textit{TiMINo} time series that avoids drawing wrong causal conclusions in the presence of confounders.
Throughout the paper, we assume the full-time graph is \textit{sufficient} (\eg there are no unobserved confounders \cite{peters2017elements}).
We will show how, even in this optimistic scenario, our negative results for temporal recourse still hold.

\begin{table*}[t]
\centering
\caption{\textbf{Overview of the theoretical results and (non-)causal recourse methods.}
We summarize the characteristics of the (non)-causal algorithmic recourse methods we considered in this work and the related theoretical results (\cref{sec:temporal-algorithmic-recourse}). 
\TSAR is the only approach robust to time (upon possessing an estimator of the stochastic process). 
\IMF is a non-causal method, so it assumes features can be manipulated \textit{independently from each other}, which is seldom the case in applications.
%
}
\label{app:tab:methods-overview}
\resizebox{0.9\textwidth}{!}{%
\begin{tabular}{@{}llll@{}}
\toprule
\textbf{Method} & \textbf{Causality} & \textbf{Recourse} & \textbf{Robust to time} \\ \midrule
\IMF (\citet{wachter2017counterfactual}) & - & Individualized & \No (by \cref{corollary:optim-intervention-time} and \cref{prop:robust-recourse-fails-over-time}) \\
\CAR (\citet{karimi2020algorithmic}) & Counterfactual & Individualized & \No (by \cref{corollary:optim-intervention-time} and \cref{prop:robust-recourse-fails-over-time}) \\
\SAR (\citet{karimi2020algorithmic}) & Interventional & Sub-population & \No (by \cref{corollary:optim-intervention-time} and \cref{prop:robust-recourse-fails-over-time}) \\
\TCAR (\cref{def:tar-definition}) & Counterfactual & Individualized & \No (by \cref{prop:impossibility-counterfactual-recourse} and \cref{corollary:counterfactual-recourse-impossible}) \\
\TSAR (\cref{def:tar-definition}) & Interventional & Sub-population & \Yes (upon having an estimator $\tilde{P}(\vX^t)$) \\
\bottomrule
\end{tabular}%
}
\end{table*}

\textbf{How do we represent user interventions in time?}
\GDT{Intuitively, the validity of recourse can only be compromised by changes occurring \textit{after} the time $t$ at which the recourse is issued, if these changes influence the distribution of the future state $\vX^{t+\tau}$.
In the unlikely scenario where recourse is framed as a hard intervention, $do(\vX^{t+\tau} = \vtheta)$, affecting \textit{all} variables $\vX$, the future state $\vX^{t+\tau}$ becomes entirely independent of the past states. This is because hard interventions \textit{detach} all variables from their causal parents, effectively overriding temporal dependencies and eliminating potential temporal effects \cite{dominguez2022adversarial}. Consequently, recourse in this case remains valid \textit{by construction}.
In practice, however, recourse:
\textit{(i)} is typically \textit{restricted to a subset of variables} $\calI$, due to constraints such as actionability or cost; and
\textit{(ii)} may be implemented as a \textit{soft intervention}, represented as $do(\vX^t_\calI = \vx^t_\calI + \vtheta)$, without fully detaching $\vX^t$ from its causal structure \cite{dominguez2022adversarial}.
In this work, we focus on this more realistic setting of soft interventions, where only a limited subset of variables is targeted, and temporal dependencies persist.
}

\GDT{
The next challenge lies in capturing the \textit{temporal nature} of actions that decision subjects can realistically perform within the causal framework. 
Consider the following example: at time $t$, Alice is denied a loan application, and the recourse action is ``get a degree.” Suppose it takes Alice $\tau$ timesteps to complete the degree. She begins studying at $t$ (\eg enrolling in university) and obtains the degree at $t + \tau$. 
The action “get a degree” represents a process initiated at time $t$ (enrolling) but whose causal effect materializes at $t + \tau$ (degree completion).
To model this, we simplify the causal framework by treating the ``enrol/get a degree'' process as a single intervention $\vtheta$ applied at $t + \tau$, assuming its causal effects are fully realized at that point.
Thus, since all user actions require time to implement, we define any action as an intervention occurring at a future $t + \tau$, where $\tau > 0$.
} 

\GDT{For the rest of the paper, in the context of temporal algorithmic recourse, we will use the  $do(\vtheta)$ notation to represent an intervention \textbf{always} applied at time $t+\tau$, unless specified otherwise.}
\GDT{
Additionally, this formulation assumes our SCM describing the time series must exhibit \textit{instantaneous effects} \cite{peters2013causal}.
As we will demonstrate in the remainder of this paper, even under this simplified intervention model, the validity of algorithmic recourse is highly sensitive to potential trends or changes in the underlying stochastic process.
}
Now, equipped with the previous definitions, we can formulate the Temporal Causal AR optimization problem as follows:

\begin{definition}[Temporal Causal AR]
\label{def:tar-definition}
Consider a stochastic process $P(\vX^t, Y^t)$, a cost function $C(\cdot, \cdot)$ (\eg the $\ell_2$ norm), a constant classifier $h$ and a \GDT{user} $\vx^{t}$ such that $h(\vx^t) < 1/2$.
We want to find the cheapest intervention $do(\vX^{t+\tau}_\calI = \vx^{t+\tau}_{\calI} + \vtheta)$
\GDT{that achieve recourse at time $t +\tau$, in expectation, assuming the causal effects of the action fully manifest at time $t +\tau$:}
\begin{equation}
    \begin{aligned}
        \vtheta^* \in \min_{\vtheta \in \bbR^d}
            & \;\; \bbE \left[C(\hat{\vx}^{t+\tau}, \vx^{t+\tau})\right]
    &\mathrm{s.t.}
        \;\; \bbE \left[h(\hat{\vx}^{t+\tau} )\right] \geq 1/2
    \end{aligned}
    \label{eqn:temporal-recourse-obj}
\end{equation}
\GDT{
where $\hat{x}^{t+\tau}$ can be sampled from either the interventional (\cref{eqn:interventional-dist-over-time}), or the counterfactual distribution (\cref{eqn:counterfactual-dist-over-time}):
\begin{equation}
    \hat{x}^{t+\tau} \sim P^{do(\vX^{t+\tau}_\calI = \vx^{t+\tau}_{\calI} + \vtheta)}(\vX^{t+\tau} \mid \vX^{t+\tau}_{nd(\calI)} = \vx^{t+\tau}_{nd(\calI)}) \label{eqn:interventional-dist-over-time}
\end{equation}
\begin{equation}    
    \hat{x}^{t+\tau} \sim P^{do(\vX^{t+\tau}_\calI = \vx^{t+\tau}_{\calI} + \vtheta);\vX^{t+\tau}=\vx^{t+\tau}}(\vX^{t+\tau}) \label{eqn:counterfactual-dist-over-time}
\end{equation}
In contrast, $\vx^{t+\tau}$ is sampled from $P(\vX^{t+\tau} \mid \vX^t = \vx^t)$, conditioned on the initial user state $\vx^t$.
}
\end{definition}

As with non-temporal causal recourse, \cref{def:tar-definition} describes both \evidenzia{temporal subpopulation causal AR} (\TSAR), for a user belonging to a specific sub-group, and individualized \evidenzia{temporal counterfactual causal AR} (\TCAR), except that now the passage of time plays a central role in the optimization problem.
%
We remark that, while practical solutions for \TSAR can be devised (see \cref{sec:method}), \TCAR is intrinsically more challenging (as we discuss in \cref{sec:counterfactual-tar}).

\textbf{Na\"ively recomputing recourse is not possible.}
\GDT{
A seemingly straightforward ``solution” to address the impact of time is to allow users to recompute recourse at time $t + \tau$, using their updated state $\vx^{t+\tau}$. However, this approach is simply not possible because the implementation of the newly suggested action may itself require additional time, $\tau' > \tau$.
For instance, even if Alice were prepared to act on the updated recourse suggestion (e.g., obtain a degree) at $t+\tau$, it is unrealistic for her to achieve this instantaneously. By the time Alice completes the action—at some later time $t+\tau'$ where $\tau' > \tau$—her state may no longer satisfy the updated requirements for recourse. In such a scenario, we might need to propose yet another action, such as ``raise yearly income by 10\%'' potentially leading to the same negative outcome.
}

\subsection{Uncertainty Over Time Compromises Counterfactual Recourse}
\label{sec:counterfactual-tar}

We begin by studying temporal counterfactual AR (\TCAR).
It provides \textit{individualized} suggestions, thus achieving the true optimal intervention for a given user \cite{karimi2020algorithmic}.
According to \cref{def:tar-definition}, it must be computed considering the \textit{future} counterfactual distribution $P^{do(\vtheta);\vX^{t+\tau}=\vx^{t+\tau}}(\vX^{t+\tau})$.
Unfortunately, this turns out to be problematic. The following proposition shows that this distribution cannot be recovered exactly except under strong assumptions.

\begin{proposition}
Let $P(\vX^t, Y^t)$ satisfy TiMINo.
Given a realization $\vx^t$ and an intervention $\vtheta \in \bbR^d$, we can recover the counterfactual distribution over the future $P^{do(\vtheta), \vX^{t+\tau}=\vx^{t+\tau}}(\vX^{t+\tau})$ if and only if, for all $t > 0$, $\mathrm{Var}(\vU^t) = 0$ and $\bbE[\vU^t]$ are constant.
\label{prop:impossibility-counterfactual-recourse}
\end{proposition}

All proofs can be found in \cref{app:proofs}.
In words, given $\vx^t$, one cannot know the true counterfactual distribution at time $t+\tau$ unless all exogenous factors have zero variance, that is $P(\vX^{t+\tau}, \ldots, \vX^{t+1} \mid \vx^t)$ is \textit{deterministic}.
\cref{prop:impossibility-counterfactual-recourse} has profound consequences for recourse because, recalling the central role of the counterfactual distribution in \cref{eqn:causal-recourse-equation}, it entails that \evidenzia{for non-deterministic processes we cannot solve \TCAR optimally}.

\begin{corollary}[Informal]
    Let $P(\vX^t)$ satisfy TiMINo \GDT{and consider a constant injective classifier $h$.}
    Given a realization $\vx^t$, a counterfactual recourse $\vtheta \in \bbR^d$ applied at time $t+\tau$, with $\tau > 0$, cannot be optimal unless exogenous factors have zero variance.
    \label{corollary:counterfactual-recourse-impossible}
\end{corollary}

We remark that \cref{prop:impossibility-counterfactual-recourse} also holds for non-TiMINo stochastic processes as long as they admit performing abduction (\eg the structural equations are invertible), and so does \cref{corollary:counterfactual-recourse-impossible}.
We explore this issue empirically in \cref{sec:experiments}.

\subsection{Sub-Population AR Deteriorates in a Non-Stationary World}
\label{sec:interventional-tar}

Given the inherent limitations of temporal counterfactual AR, in the remainder, we focus on temporal \textit{sub-population} AR (\TSAR), which is generally regarded as the most \textit{plausible} form of recourse \cite{karimi2020algorithmic}.
The next proposition shows that, insofar as $P(\vX^t, Y^t)$ is \textit{stationary}\footnote{
A discrete stochastic process $\{\vX^t\}_{t \in \bbN}$ is (weak-sense) stationary when it satisfies the following properties: $\bbE [ \vX^{t+\tau} - \vX^{t} ] = 0 $ and $ K(t+\tau, t) = K(\tau, 0) $ for all $t,\tau \in \bbN$ where $K(p,q) = \bbE\left[ (X^{p} - \bbE[X^{p}])(X^{q} - \bbE[X^{q}])\right] $ is the autocovariance and $\bbE[|\vX^t|^2] < \infty$ for all $t \in \bbN$.
} and the classifier $h$ is constant and injective, recourse that is optimal for \textit{static} sub-population recourse (\cref{eqn:causal-recourse-equation}) remains optimal over time (\cref{eqn:temporal-recourse-obj}).
\begin{proposition}
    Consider a \GDT{stationary} stochastic process $P(\vX^t)$ and a constant injective classifier $h$.  Any optimum $\vtheta^*$ of \cref{eqn:causal-recourse-equation} is also optimal for \cref{eqn:temporal-recourse-obj} for any time lag $\tau \in \bbN$.
    \label{prop:stationary-equivalence-recourse}
\end{proposition}

Despite this positive result, the issue is that \evidenzia{stationarity is seldom satisfied in practice}: many real-world processes exhibit trends (\eg inflation rate, seasonality of loan interests, \etc).
The next example shows how recourse can become invalid if the $P(\vX^t, Y^t)$ is \textit{not} stationary, even for a simple one-dimensional \textit{trend-stationary}\footnote{
A stochastic process $\{\vX^t\}_{t\in\bbN}$ is trend-stationary when it can be expressed as $\vX^t = f(t) + \ve^t$, where $f(t)$ is (non-)linear trend function and $\ve^t$ is a stationary stochastic process. 
} stochastic process. 
Full derivations are in \cref{app:proofs}.

\begin{example}
\label{example:fragility-of-recourse}
Consider a trend-stationary stochastic process defined by these structural equations:
\[
    \begin{aligned}
        & X^t = \alpha X^{t-1} + m(t) + U_X^t,
            & U_X^t \sim \calN(\mu_X,\sigma_X) \qquad
        \\
        & Y^t = \beta X^t + U_Y^t,
            & U_Y^t \sim \calN(0,1) \qquad
    \end{aligned} 
\]
for all $t$, where $\alpha \in (0, 1)$ and $\beta \in \bbR$.
The function $m(t): \bbR \rightarrow \bbR$ represents a \textit{trend} independent of $X^t$ and $Y^t$.
We consider a linear trend $m(t) = -ct + U_m^t$, where $U_m^t \sim \calN(\mu_m,\sigma_m)$ and $c \in \bbR^+$.
Consider the fixed classifier $h(X^t) = \sigma(Y^t \mid X^t)$ where $\sigma(x) = 1/(1+e^{-x})$.
We have that the optimal intervention $\theta^{t+\tau} \in \bbR$ for which we have $\bbE[h(X^{t+\tau} + \theta)] \geq 1/2$ can be expressed as:
\[
    \textstyle
    \theta^{t+\tau} = -\alpha^{\tau+1}x^{t-1} -\sum_{i=0}^{\tau} \alpha^{\tau-i} (-c(t+i) + \mu_m + \mu_X)
\]
\end{example}
Since $m(t)$ is monotonically decreasing for $c>0$, we have the optimal interventions satisfy $\theta^t \leq \theta^{t+\tau}$, implying that \evidenzia{a recourse issued at time $t$ becomes invalid as time passes}.
Following \cref{prop:stationary-equivalence-recourse}, we can state the following general corollary regarding our ability to provide optimal subpopulation recourse for general stochastic processes:
\begin{corollary}[Informal]
    Consider a discrete-time process $P(\vX^t)$ \GDT{and a constant injective classifier $h$}.  Unless $P(\vX^t)$ is stationary, the optimal intervention $\vtheta^*$ achieving recourse can vary depending on $t, \tau \in \bbN$.
    \label{corollary:optim-intervention-time}
\end{corollary}

\subsection{Robust Algorithmic Recourse Is Not Enough To Counteract Time}
\label{sec:robust-tar}

Given the previous results, we could imagine \textit{robustifying} the recourse procedure to account for non-stationarity of $P(\vX^t, Y^t)$. 
For example, a common solution to robustify \CAR and \SAR is to provide an intervention $\vtheta$ achieving recourse within a causal uncertainty set $B(\vX; \Delta)$ \cite{dominguez2022adversarial}.
In this section, we show how \evidenzia{set-based robust causal recourse method falls short when dealing with time}.
We first start by defining robust causal AR:
\begin{definition}[Adapted from \citet{dominguez2022adversarial}]
Consider a realization $\vx \in \bbR^d$, a norm $||\cdot||$, and a tolerance $\epsilon > 0$. We define a causal uncertainty set $B(\vx; \vDelta) = \{ \vx' \sim P^{do(\vDelta), \vX=\vx}(\vX) : || \vDelta || \leq \epsilon \}$ as the collection of the causal counterfactuals under small additive \GDT{perturbations} $\vDelta \in \bbR^d$.
We want to find the cost-minimizing intervention $\vtheta$ achieving recourse in \textit{all} the region defined by $B(\vx; \vDelta)$.
Thus, the optimization objective for robust recourse becomes:
\begin{equation}
    \begin{aligned}
        \vtheta^* \in \argmin_{\vtheta \in \bbR^d} \ \bbE [C(\hat{\vx}, \vx)] \quad \mathrm{s.t.} \;\; \bbE[h(\hat{\vx})] \geq 1/2 \;\; \forall \; \hat{\vx} \sim B(\vx; \vDelta)
    \end{aligned}
    \label{eqn:robust-causal-recourse}
\end{equation}
where $\hat{\vx}$ is distributed according to either the counterfactual or interventional distribution.
\label{def:robust-causal-recourse}
\end{definition}

A robust intervention might still obtain recourse for later time steps depending on the tolerance $\epsilon$. Intuitively, by asking the user to perform a more difficult action (\eg increase your income by $\$1000$, instead of $\$100$), we can provide interventions that are less susceptible to potential dynamics. However, if the intervention is applied too late, we will not achieve recourse, as shown by the following proposition:

\begin{proposition}
    Consider a fixed $\epsilon > 0$, a trend-stationary process $P(\vX^t)$, a constant injective classifier $h$ and realization $\vx^t$ where $h(\vx^t) < 1/2$. Let us assume we have an optimal $\epsilon$-robust intervention $\vtheta$ for timestep $t$. There always exists a trend $m: \bbN \rightarrow \bbR$ and a positive $\tau$, such that $\bbE_{\vx^{t+\tau} \sim P^{do(\vtheta)}(\vX^{t+\tau} \mid \vX_{nd(\calI)}^{t+\tau} = \vx_{nd(\calI)}^{t+\tau}, \vX^t = \vx^t)}{[h(\vx^{t+\tau})]} < 1/2$.
    \label{prop:robust-recourse-fails-over-time}
\end{proposition}

\subsection{On the Stability of Recourse Over Time}
\label{sec:stability-of-tar}

In \cref{sec:counterfactual-tar,sec:interventional-tar,sec:robust-tar}, we showed how recourse validity can be compromised by the uncertainty and non-stationarity of the stochastic process $P(\vX^t, Y^t)$, and we also showed how set-based robustness techniques fail over time. 
However, \GDT{decision subjects} might be willing to accept recourses that slowly become less effective rather than performing more challenging interventions.
Thus, we now characterize instead the \textit{rate} at which our recourse suggestion validity decreases, \GDT{while assuming a non-stationary $P(Y^t \mid \vX^t)$ approximated by a sequence of classifiers $h^t$, one for each $t \in \bbN$.}

\begin{definition}[Temporal recourse invalidation rate]
Consider a discrete-time stochastic process $P(\vX^t, Y^t)$, and any classifier $h^t$ approximating $P(Y^t \mid \vX^t)$ for each $t \in \bbN$.
Given a realization $\vx^t$ and an intervention $\vtheta$ such that $\bbE[h(\hat{\vx}^t)] \geq 1/2$, where $\hat{\vx}^t \sim P^{do(\vtheta)}(\vX^t \mid \vX_{nd(\calI)}^t = \vx_{nd(\calI)}^t)$, we define the temporal invalidation rate after a time-lag $\tau > 0$ as:
\begin{equation}
\label{eq:inv_rate}
   \Delta h(\vtheta; \tau) = \bbE\left[\left|h^{t+\tau}(\hat{\vx}^{t + \tau})-h^{t}(\hat{\vx}^t)\right|\right]
\end{equation}
where $\hat{\vx}^{t+\tau} \sim P^{do(\vtheta)}(\vX^{t+\tau} \mid \vX_{nd(\calI)}^{t+\tau} = \vx_{nd(\calI)}^{t+\tau})$.
\end{definition}

Let us now consider the setting in which we have a bounded stochastic process $P(\vX^t, Y^t)$, where $-k \leq X^t_i \leq k$ for some $k \in \bbR^+$ for all $t \in \bbN$. 
If we have access to a dataset $\calD^t \sim P(\vX^t, Y^t)$, we can use it to train a classifier $h^t$ via empirical risk minimization.
Let us consider a \textit{linear} classifier $h^t(\vx) = \dprod{\vbeta^t}{\vx^t}$ with bounded weights $-k \leq \beta^t_i \leq k$ \eg trained via Bounded Least-Squares (BLS) \cite{stark1995bounded}. 
Then, given an intervention $\vtheta$, we can derive the following upper bound on the recourse instability within a time interval $(t, t+\tau)$.

\begin{theorem}[Upper-bound invalidation rate]
    Consider a dis\-cre\-te-time stochastic process $P(\vX^t, Y^t)$ and a sequence of linear classifiers $h^t(\vx^t) = \dprod{\vbeta^t}{\vx^t}$ approximating $P(Y^t \mid \vX^t)$.
    We assume $-k \leq \beta_i, X_i \leq k$ for $k \in \bbR^+$.
    The temporal invalidation rate is upper-bounded as follows:
    \begin{equation}
        \Delta h(\vtheta; \tau) \leq k\sqrt{d} \cdot \expect{ \norm{\vbeta^{t+\tau}-\vbeta^t}}+ \expect{ \norm{\hat{\vx}^{t+\tau}-\hat{\vx}^t}} 
    \end{equation}
    where $\norm{\cdot}$ is the $\ell_2$-norm and the expectation is over $\calD^t \sim P(\vX^t, Y^t)$ and the training process. 
    \label{theorem:invalidation-rate}
\end{theorem}
\cref{theorem:invalidation-rate} shows how the recourse instability is upper bounded by how much the world varies between $t$ and $t+\tau$ in terms of the data distribution $P(\vX^t)$, and the classifier.
Moreover, the size of the problem $d$ concurs by increasing the worst-case error at a sublinear rate.
\GDT{The upper bound can be useful for non-linear classifiers $h$ if we consider a linear function approximating locally \cite{simonyan2013deep} their decision function close to a realization $\vx^t$, such as LIME \cite{ribeiro2016should}.}
If our stochastic process is \textit{trend-stationary} as in \cref{example:fragility-of-recourse}, we can derive the following upper bound:
\begin{corollary}
    Consider a discrete-time trend-stationary stochastic process $P(\vX^t)$, where $m_i: \bbN \rightarrow \bbR$ represents the trend for $X_i$ \GDT{and the classifiers $h^t(\vx^t) = \dprod{\vbeta^t}{\vx^t}$.}
    Let us define $m^*(t) = \max_{i \in [d]} m_i(t)$ as the largest trend for $t \in \bbN$.
    Then, we have the upper bound:
    \begin{equation}
            \Delta h(\vtheta; \tau) \leq k\left(\sqrt{d} \expect{ \norm{\vbeta^{t+\tau}-\vbeta^t}} + d\left(m^*(t+\tau)-m^*(t)\right)\right) 
    \end{equation}
    \label{corollary:invalidation-rate-trend}
\end{corollary}
These results assume that the cost function remains constant over time.
In \cref{app:additional_result_cost_stability}, we show an analogous result when this is not the case, as in \textit{personalized cost functions} \cite{detoni2024personalized,esfahani2024preference}.

\subsection{Accounting for Time in Practice}
\label{sec:method}
In \cref{sec:counterfactual-tar,sec:interventional-tar,sec:robust-tar,sec:stability-of-tar}, we showed how recourse validity is hindered by the \textit{uncertainty} and \textit{non-stationarity} of the stochastic process.
Given a factual instance $\vx$ and a recourse $\vtheta$, \cref{theorem:invalidation-rate} also implies the upper bound can grow quickly depending on the difference of the induced interventional distributions and classifier $h$ between $t$ and $t+\tau$.
Unfortunately, for any model $h$ and tolerance $\epsilon$, there can exist \textit{multiple} (robust) recourses $\vtheta$ to choose from \cite{pawelczyk20modelmulti}.
Since (robust) \CAR and \SAR have \textit{no means to differentiate between recourses}, they might end up suggesting interventions which rapidly become invalid.
An alternative is to settle for classical robust AR, which \textit{can} provide some amount of safety w.r.t. time depending on the chosen epsilon $\epsilon$ (cf. \cref{sec:experiments}).  The issue with this is that, as shown by \cref{prop:robust-recourse-fails-over-time}, choosing $\epsilon$ without considering how the world changes can be dramatically suboptimal, \ie robust AR might recommend expensive actions that risk becoming invalid.

Luckily, for \TSAR, we can mitigate these issues, as long as we have access to an estimator of the stochastic process.
We present a simple algorithm (\cref{alg:differentiable-temporal-recourse}) for temporal \textit{sub-population} algorithmic recourse (\cref{eqn:temporal-recourse-obj}) drawing inspiration from \textit{adversarially robust recourse} \cite{dominguez2022adversarial}.
Following the results of \cref{sec:robust-tar}, we argue that, instead of providing robust recourse for an arbitrary uncertainty set with a fixed $\epsilon$, we need to provide a robust $\vtheta$ for a \textit{forecasted} region of the feature space.
We do so by extending the notion of uncertainty set $B(\vx^t; \tau)$ to consider the distribution \GDT{entailed by the TiMINo SCM} conditioned on the observed realization, $P(\vX^{t+\tau} \mid \vX^t = \vx^t)$, after a time lag $\tau > 0$:
\begin{equation}
     B(\vx^t; \tau) = \{ \vx' \sim P(\vX^{t+\tau} \mid \vX^t = \vx^t) \}
     \label{eqn:temporal-uncertainty-set}
\end{equation}
Such a region does not depend on a fixed $\epsilon$, thus sidestepping the issue shown by \cref{prop:robust-recourse-fails-over-time}. 
\cref{alg:differentiable-temporal-recourse} assumes to have access to a constant and \textit{differentiable} classifier $h$, and to an estimator $\tilde{P}(\vX^t)$ of the stochastic process.
We define $\mathrm{ER}(\vx, \tau; \vtheta) = \bbE[h(\hat{\vx})]$ where $\hat{\vx}$ is sampled from the interventional distribution conditioned on the non-descendant $nd(\calI)$ of the intervened upon nodes $\calI$.
Similarly to \citet{dominguez2022adversarial}, we approximate $B(\vx^t; \tau)$ by sampling a finite number of instances from $P(\vX^{t+\tau} \mid \vX^t = \vx^t)$.
As usual, users can control the trade-off between cost and robustness by varying the $\lambda$ factor (line 5, \cref{alg:differentiable-temporal-recourse}).
\GDT{
In practice, to solve the optimization problem in \cref{def:tar-definition}, we follow the causal recourse approach of \citet{dominguez2022adversarial} and use projected gradient descent over the (feasible) actions and projected gradient ascent over the temporal uncertainty set (line 6 and 4 \cref{alg:differentiable-temporal-recourse}, respectively).
}

\textbf{Computational complexity.} The running time of \cref{alg:differentiable-temporal-recourse} depends on (i) the number of epochs $N$, and (ii) an upper bound on the number of iterations $K$ of the inner loop (lines $3$-$6$).
Considering all potential variable subsets $\calI \subset [d]$, the complexity is $O(NK 2^{d})$.
Luckily, not all features are actionable, and the general wisdom is to provide sparse solutions (\eg by considering only $|\calI| \leq m$ sets) so we have $O(N K\binom{d}{m})$ if $m \ll d$.
Lastly, the time-lag $\tau$ has no impact on the running time of the algorithm, but we would need to run \cref{alg:differentiable-temporal-recourse} for each $\tau$ specified by the user. 

\textbf{Relationship between \cref{alg:differentiable-temporal-recourse} and other causal methods.}
\cref{alg:differentiable-temporal-recourse} subsumes existing causal recourse methods.
If we replace $B(\vx^t; \tau)$ with the uncertainty set in \cref{def:robust-causal-recourse}, we obtain the \textit{robust} counterfactual (\CAR) or sub-population (\SAR) recourse method \cite{dominguez2022adversarial}, depending on how we define the distribution over the expectation in $\mathrm{ER}(\vx, \tau; \vtheta)$.
Moreover, if we \GDT{do not consider $\tau$ such as $\mathrm{ER}(\vx; \vtheta) = h(\vx + \vtheta)$ and $B(\vx^t; \vDelta) = \{\vx^t + \vDelta : || \vDelta || \leq \epsilon \}$, we obtain (robust) non-causal recourse (\IMF, \cite{wachter2017counterfactual}).}

\begin{algorithm}[t]
\caption{Generate robust recourse solutions for \TSAR given a future time-lag $\tau$, $\gamma < 1$, $\lambda > 1$, a differentiable classifier $h$, an estimator $\tilde{P}(\vX^t)$ and a subset of intervened nodes $\calI \subseteq [d]$. We denote with $\ell$ the \textit{binary cross-entropy loss}.}
\label{alg:cap}
\begin{algorithmic}[1]
\Require $\vx^t$, individual; $N > 0$; $\lambda > 0$; $\eta > 0$; 
\State $B(\vx^t; \tau) \gets \{ \vx' \sim \tilde{P}(\vX^{t+\tau} \mid \vX^t = \vx^t) \}$
\For{$epochs = 1$ to $N$}
\While{$\exists \; \vx' \in B(\vx^t; \tau) $ s.t. $ \mathrm{ER}(\vx', \tau; \vtheta) < 1/2 $}
\If{it converges}
\State \Return $\vtheta$
\EndIf
\State $\vx^* \gets \argmax_{\vx' \in B(\vx^t; \tau)} \ell(\mathrm{ER}(\vx', \tau; \vtheta), \bm{1})$
\State $\calL \gets \ell(\mathrm{ER}(\vx^*, \tau; \vtheta), \bm{1}) + \lambda\| \vtheta \|$ 
\State $\vtheta \gets \vtheta - \eta \nabla \calL$
\EndWhile
\State $\lambda \gets \gamma\lambda$
\EndFor
\State \Return $\vtheta$
\end{algorithmic}
\label{alg:differentiable-temporal-recourse}
\end{algorithm}

\section{Empirical Evaluation}
\label{sec:experiments}

In this section, we empirically study the effect of time on recourse validity in synthetic and realistic settings taken from the literature, by comparing \cref{alg:cap} against several robust (non-)causal AR methods.
The source code, datasets, and raw results are available on GitHub under a permissive license\footnote{\url{https://github.com/unitn-sml/temporal-algorithmic-recourse}}. 
See \cref{app:implementation_details_experiments} for a detailed explanation of the experimental setting, techniques, and training setup. 

\subsection{Experiments With Synthetic Time-Series}
\label{sec:synthetic-experiment}

\textbf{Experimental setup.} First, we consider the linear and non-linear 3-variable synthetic ANMs from \citet{karimi2021algorithmic} representing a binary decision problem (\eg loan granted/denied).
We adapt them to describe a trend-stationary stochastic process by adding an additive trend function $m(t) = \alpha \cdot (\beta_l \cdot l(t) + \beta_s \cdot s(t))$ to the structural equations, where $l(t)$ and $s(t)$ are the \textit{linear} and \textit{seasonal} components.
The parameter $\alpha \in (0,1)$ governs the strength of the trend.
We consider three types of trends: \textit{linear} ($\beta_l >0, \beta_s = 0$), \textit{seasonal} ($\beta_l = 0, \beta_s > 0$) and \textit{linear+seasonal} ($\beta_l > 0, \beta_s > 0$).
Then, we sample a time series for each ANM with 10000 individuals for $t \in [0, 100]$ timesteps.
We split the time series into training (8000) and testing (2000) and train a fixed 3-layer MLP to approximate $P(Y^t \mid \vX^t)$ using \textit{only the training data at time $t=0$}.
We pick 250 individuals negatively classified ($h(\vx) < 1/2$) by the MLP from the test set at time $t=0$, and we compute recourse suggestions, by considering the $\ell_1$-norm as a cost function, with \textit{robust} counterfactual and sub-population recourse (\CAR and \SAR, \cite{dominguez2022adversarial}), \textit{robust} non-causal recourse (\IMF, \cite{wachter2017counterfactual}) and time-aware sub-population recourse (\TSAR, \cref{alg:differentiable-temporal-recourse}).
We empirically choose a smaller and larger $\epsilon \in \{3,5\}$ maximizing the robust methods' validity at $t=0$.
To simulate the user implementing the suggested intervention at a later time, we vary the time lag $\tau$ and compute the \evidenzia{empirical average validity} (\% of interventions achieving recourse) for each method at time $\tau$.
We repeat the procedure 10 times.
In these experiments, we assume to know the true causal graph and structural equations.

\begin{figure}
    \centering 
    \includegraphics[width=0.5\linewidth]{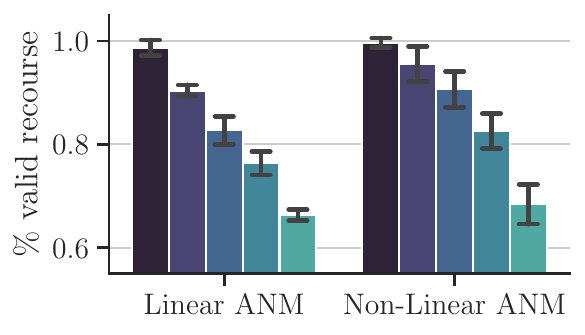}
    \caption{\textbf{Effect of uncertainty on counterfactual AR.}
    Empirical average validity and standard deviation over 10 runs of \textit{robust} counterfactual algorithmic recourse (\CAR) at $t=50$. We vary the variance $\sigma_U$ of the exogenous factors of the stochastic process.
    Legend ($\sigma_U$): 
    \raisebox{0.5ex}{\fcolorbox[HTML]{FFFFFF}{2E1E3B}{\rule{0pt}{1pt}\rule{1pt}{0pt}}} $0$
    \raisebox{0.5ex}{\fcolorbox[HTML]{FFFFFF}{413D7B}{\rule{0pt}{1pt}\rule{1pt}{0pt}}} $0.3$
    \raisebox{0.5ex}{\fcolorbox[HTML]{FFFFFF}{37659E}{\rule{0pt}{1pt}\rule{1pt}{0pt}}} $0.5$
    \raisebox{0.5ex}{\fcolorbox[HTML]{FFFFFF}{348FA7}{\rule{0pt}{1pt}\rule{1pt}{0pt}}} $0.7$
    \raisebox{0.5ex}{\fcolorbox[HTML]{FFFFFF}{40B7AD}{\rule{0pt}{1pt}\rule{1pt}{0pt}}} $1.0$.
    }
    \label{fig:uncertainty-counterfactual-recourse}
\end{figure}

\begin{figure*}
\centering
\begin{minipage}{\textwidth}
\centering
\includegraphics[width=\linewidth]{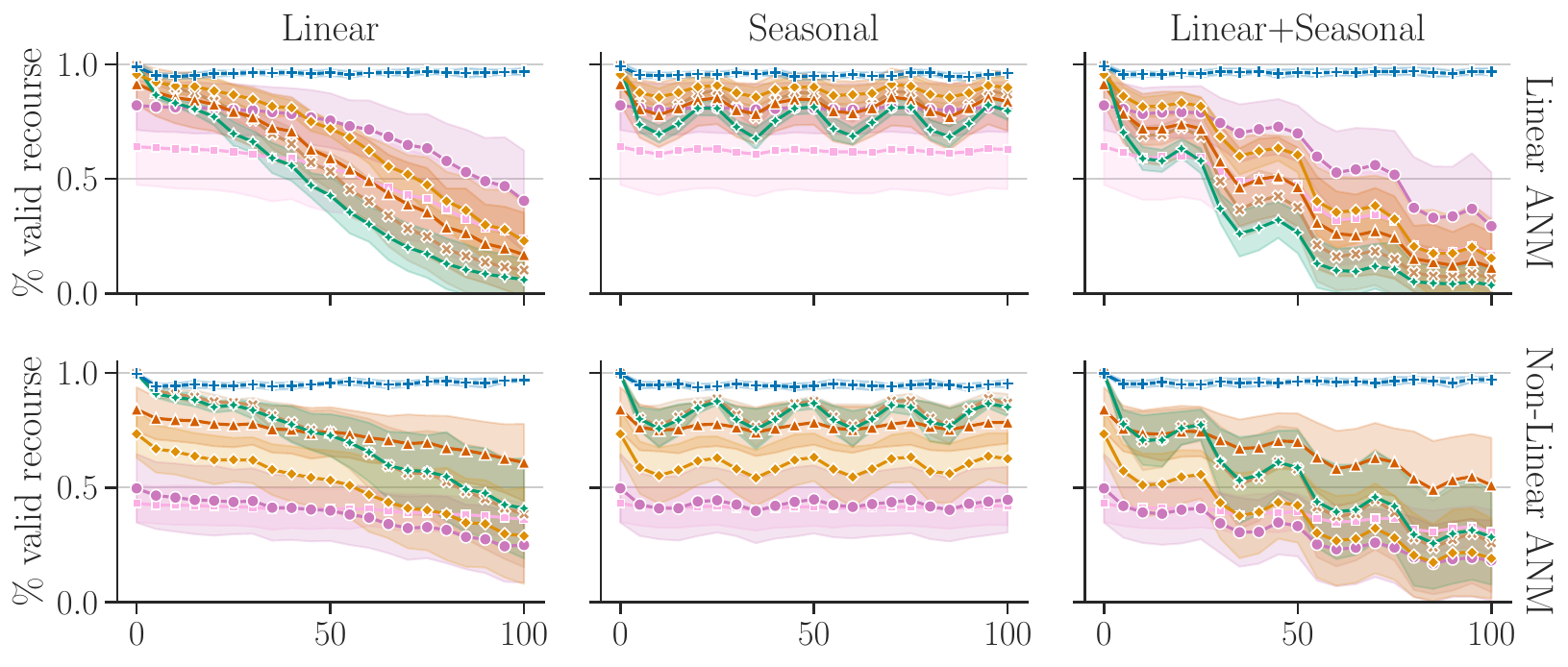}
\end{minipage}

\caption{
\textbf{Causal algorithmic recourse on diverse time series.}
Empirical average validity and standard deviation (10 runs) for the robust  ($\epsilon \in \{3, 5\}$) and time-aware causal recourse methods for the synthetic ANMs under different trends ($\alpha=1.0$).
Legend:
    \raisebox{0.5ex}{\fcolorbox[HTML]{FFFFFF}{0173B2}{\rule{0pt}{1pt}\rule{5pt}{0pt}}} \TSAR
    \raisebox{0.5ex}{\fcolorbox[HTML]{FFFFFF}{DE8F05}{\rule{0pt}{1pt}\rule{5pt}{0pt}}} \CAR ($\epsilon = 3$)
    \raisebox{0.5ex}{\fcolorbox[HTML]{FFFFFF}{029E73}{\rule{0pt}{1pt}\rule{5pt}{0pt}}} \SAR ($\epsilon = 3$)
    \raisebox{0.5ex}{\fcolorbox[HTML]{FFFFFF}{D55E00}{\rule{0pt}{1pt}\rule{5pt}{0pt}}} \IMF ($\epsilon = 3$)
     and 
    \raisebox{0.5ex}{\fcolorbox[HTML]{FFFFFF}{CC78BC}{\rule{0pt}{1pt}\rule{5pt}{0pt}}} \CAR ($\epsilon = 5$)
    \raisebox{0.5ex}{\fcolorbox[HTML]{FFFFFF}{CA9161}{\rule{0pt}{1pt}\rule{5pt}{0pt}}} \SAR ($\epsilon = 5$)
    \raisebox{0.5ex}{\fcolorbox[HTML]{FFFFFF}{FBAFE4}{\rule{0pt}{1pt}\rule{5pt}{0pt}}} \IMF ($\epsilon = 5$).
}
\label{fig:results-synthetic-data}
\end{figure*}

\textbf{Uncertainty invalidates counterfactual recourse (\CAR) over time.}
First, we consider a \textit{stationary} version of the synthetic ANMs with exogenous noise $U_i \sim \calN(0, \sigma_U)$ for all $i \in [3]$ and vary the variance of the exogenous factors $\sigma_U \in \{0, 0.3, 0.5, 0.7, 1.0\}$, where $\sigma_U=0$ means that the ANM is deterministic.
For each value of $\sigma_U$, we compute the recourse suggestions using robust counterfactual recourse (\CAR).
\cref{fig:uncertainty-counterfactual-recourse} displays how validity decreases as variance increases 
showing that the validity over time of (robust) \CAR recommendations is strongly impacted by the exogenous noise, as per \cref{prop:impossibility-counterfactual-recourse}, even in an ideal case in which the SCM is stationary and known.
\GDT{\cref{app:extended-empirical-results-car} shows extended results for $t \in \{0, 100\}$.}

\textbf{Incorporating time is beneficial in causal algorithmic recourse.}
\cref{fig:results-synthetic-data} shows how \TSAR (\cref{alg:differentiable-temporal-recourse}) achieves superior validity over time than robust (non-)causal methods on the synthetic settings, considering the diverse type of trends $m(t) \in \{\hbox{Linear}, \hbox{Seasonal}, \hbox{Linear+Seasonal} \}$.
Interestingly, robust causal recourse methods depend highly on the chosen hyperparameters ($\epsilon$, $\eta$, and $\lambda$) while \TSAR requires less tuning.
For example, both \CAR and \IMF show worse validity on the non-linear ANM when increasing $\epsilon$.
\GDT{Lastly, in \cref{app:sec:trade-off-validity-cost,app:futher_analysis_cost_sparsity} we provide further experiments on the \textit{tradeoff} between \textit{validity over time} and \textit{cost}, and on the \textit{sparsity} of recourses, respectively.}%

\subsection{Experiments With Realistic Time-Series}
\label{sec:real-experiment}

In \cref{sec:synthetic-experiment}, we assume perfect knowledge of the causal graph and structural equations governing the stochastic process. 
We now relax these assumptions by learning the structural equations in a data-driven manner, using a simple generative model, on three datasets. 

\textbf{Experiments setup}.
We consider three real-world datasets concerning high-risk decision tasks: recidivism prediction (\texttt{COMPAS} \cite{angwin2016machine}), and loan approval (\texttt{Adult} \cite{Dua:2019} and \texttt{Loan} \cite{karimi2021algorithmic}).
They involve categorical and continuous features, some of which are not actionable (\eg age, ethnicity, etc.). 
We use the causal graphs defined by \citet{nabi2018fair} for \texttt{COMPAS} and \texttt{Adult}, and \citet{karimi2021algorithmic} for \texttt{Loan}. 
We extend these datasets by adding linear+seasonal trends simulating real-world phenomena (\eg income can fluctuate depending on the job market or individual expenses). 
Full details are available in \cref{app:implementation_details_experiments}.
Given the ground truth SCMs, we sample an additional separate time series with 2000 individuals for $t \in [0,100]$, and we use all the samples up to $t=50$ to learn an approximate SCM for each real-world dataset.
We approximate the structural equations using a CVAE-like generative model \cite{sohn2015learning}.
In the experiment, all methods use the same approximate SCMs to compute recourse.
Lastly, we perform the same evaluation procedure used for the synthetic experiments.

\textbf{Incorporating time is beneficial also with approximate SCMs.}
\GDT{\cref{tab:realistic_dataset}} shows how \TSAR \GDT{can provide more robust} recourse recommendations than the counterparts by exploiting an estimator $\tilde{P}(\vX^t)$.
In both {\tt Adult} and {\tt Loan}, \TSAR shows \GDT{comparable} or better validity than the non-temporal methods. 
For example, in {\tt Loan}, \TSAR achieve almost twice the \GDT{expected} average validity ($\sim 72\%$) of the best non-temporal approach \SAR ($\sim 39\%$) for $t=100$, \GDT{albeit exhibiting larger variance.}
Understandably, \TSAR's performance hinges on the quality of the underlying estimator.
In {\tt COMPAS}, while \TSAR has good performance for $t \in \{50, 60, 70\}$, it gracefully degrades afterwards. This occurs because the trend estimator underestimates the trend impact on the features.
In fact, \cref{app:experimental-setting-real} shows that, if we employ a perfect estimator $P(\vX^t)$, \TSAR outperforms all non-temporal methods (\cref{app:fig:task-3-ground-truth}).
As in \cref{sec:synthetic-experiment}, the non-temporal methods are sensitive to the hyperparameters, \eg for {\tt COMPAS}, \CAR provides higher validity for $\epsilon=0.05$ rather than $\epsilon=0.5$, while \TSAR needs less tuning.

\begin{table*}[t]
\centering
\caption{\textbf{Effect of time on realistic datasets.} Empirical average validity and standard deviation (10 runs) for the robust  ($\epsilon \in \{0.05, 0.5\}$) and time-aware causal recourse methods for the realistic datasets under a non-linear trend for $t \in \{50, 60, 70, 100\}$.
We denote the best results in bold with a shaded background.
}
\resizebox{0.9\textwidth}{!}{%
\begin{tabular}{ccccccccc}
\toprule
\multirow{2}{*}{} & \multirow{2}{*}{$t$} & \multicolumn{2}{c}{\texttt{CAR}} & \multicolumn{2}{c}{\texttt{SAR}} & \multicolumn{2}{c}{\texttt{IMF}} & {\texttt{T-SAR}} \\
 &  & $\epsilon = 0.05$ & $\epsilon = 0.5$ & $\epsilon = 0.05$ & $\epsilon = 0.5$ & $\epsilon = 0.05$ & $\epsilon = 0.5$ & (\cref{alg:differentiable-temporal-recourse})  \\ \midrule
 & 50 & $0.86 \pm \scriptstyle 0.05$ & $0.89 \pm \scriptstyle 0.02$ & $0.98 \pm \scriptstyle 0.05$ & $0.54 \pm \scriptstyle 0.33$ & \cellcolor[HTML]{EFEFEF}$\bm{0.99 \pm \scriptstyle 0.02}$ & \cellcolor[HTML]{EFEFEF}$\bm{0.99 \pm \scriptstyle 0.03}$ & $0.98 \pm \scriptstyle 0.01$ \\
 & 60 & $0.77 \pm \scriptstyle 0.07$ & $0.86 \pm \scriptstyle 0.04$ & \cellcolor[HTML]{EFEFEF}$\bm{0.97 \pm \scriptstyle 0.05}$ & $0.54 \pm \scriptstyle 0.33$ & $0.88 \pm \scriptstyle 0.02$ & $0.94 \pm \scriptstyle 0.02$ & $0.95 \pm \scriptstyle 0.02$ \\
 & 70 & $0.78 \pm \scriptstyle 0.08$ & $0.86 \pm \scriptstyle 0.04$ & \cellcolor[HTML]{EFEFEF}$\bm{0.97 \pm \scriptstyle 0.05}$ & $0.54 \pm \scriptstyle 0.32$ & $0.87 \pm \scriptstyle 0.03$ & $0.94 \pm \scriptstyle 0.02$ & $0.95 \pm \scriptstyle 0.02$ \\
\multirow{-4}{*}{\rotatebox[origin=c]{90}{\texttt{Adult}}} & 100 & $0.78 \pm \scriptstyle 0.08$ & $0.86 \pm \scriptstyle 0.05$ & \cellcolor[HTML]{EFEFEF}$\bm{0.97 \pm \scriptstyle 0.05}$ & $0.54 \pm \scriptstyle 0.33$ & $0.88 \pm \scriptstyle 0.03$ & $0.93 \pm \scriptstyle 0.03$ & $0.96 \pm \scriptstyle 0.02$ \\ \midrule
 & 50 & \cellcolor[HTML]{EFEFEF}$\bm{1.00 \pm \scriptstyle 0.00}$ & $0.65 \pm \scriptstyle 0.35$ & \cellcolor[HTML]{EFEFEF}$\bm{1.00 \pm \scriptstyle 0.00}$ & $0.80 \pm \scriptstyle 0.39$ & \cellcolor[HTML]{EFEFEF}$\bm{1.00 \pm \scriptstyle 0.00}$ & \cellcolor[HTML]{EFEFEF}$\bm{1.00 \pm \scriptstyle 0.00}$ & \cellcolor[HTML]{EFEFEF}$\bm{1.00 \pm \scriptstyle 0.00}$ \\
 & 60 & $0.85 \pm \scriptstyle 0.01$ & $0.64 \pm \scriptstyle 0.35$ & $0.94 \pm \scriptstyle 0.01$ & $0.80 \pm \scriptstyle 0.39$ & $0.56 \pm \scriptstyle 0.01$ & $0.64 \pm \scriptstyle 0.01$ & \cellcolor[HTML]{EFEFEF}$\bm{1.00 \pm \scriptstyle 0.00}$ \\
 & 70 & $0.91 \pm \scriptstyle 0.01$ & $0.65 \pm \scriptstyle 0.35$ & $0.96 \pm \scriptstyle 0.01$ & $0.80 \pm \scriptstyle 0.39$ & $0.75 \pm \scriptstyle 0.01$ & $0.80 \pm \scriptstyle 0.01$ & \cellcolor[HTML]{EFEFEF}$\bm{0.99 \pm \scriptstyle 0.02}$ \\
\multirow{-4}{*}{\rotatebox[origin=c]{90}{\texttt{COMPAS}}} & 100 & $0.74 \pm \scriptstyle 0.02$ & $0.63 \pm \scriptstyle 0.34$ & \cellcolor[HTML]{EFEFEF}$\bm{0.87 \pm \scriptstyle 0.02}$ & $0.79 \pm \scriptstyle 0.39$ & $0.43 \pm \scriptstyle 0.01$ & $0.51 \pm \scriptstyle 0.01$ & $0.78 \pm \scriptstyle 0.17$ \\ \midrule
 & 50 & $0.86 \pm \scriptstyle 0.11$ & \cellcolor[HTML]{EFEFEF}$\bm{0.92 \pm \scriptstyle 0.07}$ & \cellcolor[HTML]{EFEFEF}$\bm{0.92 \pm \scriptstyle 0.05}$ & $0.78 \pm \scriptstyle 0.24$ & \cellcolor[HTML]{EFEFEF}$\bm{0.92 \pm \scriptstyle 0.07}$ & $0.91 \pm \scriptstyle 0.08$ & $0.82 \pm \scriptstyle 0.06$ \\
 & 60 & $0.39 \pm \scriptstyle 0.02$ & $0.65 \pm \scriptstyle 0.07$ & $0.54 \pm \scriptstyle 0.12$ & $0.60 \pm \scriptstyle 0.21$ & $0.46 \pm \scriptstyle 0.07$ & $0.54 \pm \scriptstyle 0.08$ & \cellcolor[HTML]{EFEFEF}$\bm{0.89 \pm \scriptstyle 0.20}$ \\
 & 70 & $0.42 \pm \scriptstyle 0.02$ & $0.66 \pm \scriptstyle 0.06$ & $0.54 \pm \scriptstyle 0.13$ & $0.61 \pm \scriptstyle 0.21$ & $0.50 \pm \scriptstyle 0.08$ & $0.57 \pm \scriptstyle 0.07$ & \cellcolor[HTML]{EFEFEF}$\bm{0.84 \pm \scriptstyle 0.29}$ \\
\multirow{-4}{*}{\rotatebox[origin=c]{90}{\texttt{Loan}}} & 100 & $0.07 \pm \scriptstyle 0.03$ & $0.24 \pm \scriptstyle 0.08$ & $0.15 \pm \scriptstyle 0.09$ & $0.39 \pm \scriptstyle 0.25$ & $0.13 \pm \scriptstyle 0.06$ & $0.17 \pm \scriptstyle 0.07$ & \cellcolor[HTML]{EFEFEF}$\bm{0.72 \pm \scriptstyle 0.41}$ \\ \bottomrule
\end{tabular}%
}
\label{tab:realistic_dataset}
\end{table*}

\begin{figure*}[t]
\centering
\begin{minipage}{\textwidth}
\includegraphics[width=\linewidth]{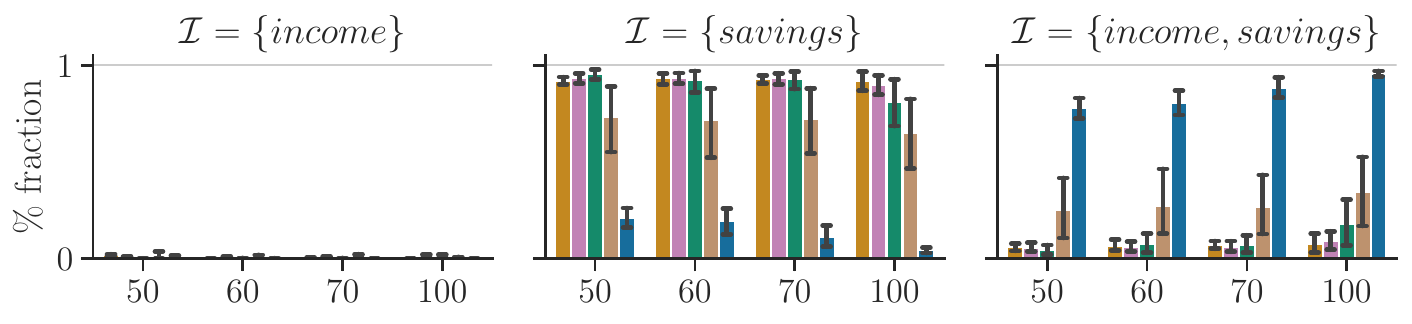}
\end{minipage}

\caption{\textbf{Effect of time on actionable features}. Distribution of the intervention sets $\calI$ over the actionable features achieving recourse on \texttt{Loan} for different $t$.
Legend:
    \raisebox{0.5ex}{\fcolorbox[HTML]{FFFFFF}{0173B2}{\rule{0pt}{1pt}\rule{5pt}{0pt}}} \TSAR
    \raisebox{0.5ex}{\fcolorbox[HTML]{FFFFFF}{DE8F05}{\rule{0pt}{1pt}\rule{5pt}{0pt}}} \CAR ($\epsilon = 0.05$)
    \raisebox{0.5ex}{\fcolorbox[HTML]{FFFFFF}{029E73}{\rule{0pt}{1pt}\rule{5pt}{0pt}}} \SAR ($\epsilon = 0.05$)
    and 
    \raisebox{0.5ex}{\fcolorbox[HTML]{FFFFFF}{CC78BC}{\rule{0pt}{1pt}\rule{5pt}{0pt}}} \CAR ($\epsilon = 0.5$)
    \raisebox{0.5ex}{\fcolorbox[HTML]{FFFFFF}{CA9161}{\rule{0pt}{1pt}\rule{5pt}{0pt}}} \SAR ($\epsilon = 0.5$)
}
\label{fig:results-real-data}
\end{figure*}

\textbf{Accounting for time ensures more targeted interventions.}
Lastly, \cref{fig:results-real-data} shows how \TSAR suggests interventions \GDT{on feature sets $\calI$} counteracting the effect of the trend more effectively than non-temporal methods.
We excluded {\tt COMPAS} from the analysis since it has a single actionable feature.
In {\tt Loan}, we have two actionable features $\{\textit{income},\textit{savings}\}$, with \textit{income} subject to a trend. \TSAR provides recourse including the trend variable, while the other methods exclude it from the recommendation, thus yielding a lower validity.
In {\tt Adult}, we again have two actionable features $\{\textit{education},\textit{work-hours-per-week}\}$, where \textit{work-hours-per-week} is subject to a trend. In this case, \TSAR suggests acting on \{\textit{work-hours-per-week}\} only. Robust sub-population methods (\SAR) will instead ask the user to act on both $\{\textit{education},\textit{work-hours-per-week}\}$ because they have to robustify on both variables since they cannot forecast how they will change.
Non-causal methods (\IMF) act on all actionable features but achieve a lower validity
since they cannot account for trend effects.%

\section{Discussion and Limitations}
We now discuss some limitations of our work, which open up interesting avenues for future work. 

\textbf{Feasibility of temporal recourse and its evaluation.}
\GDT{\cref{alg:differentiable-temporal-recourse} depends on the quality of the estimator $\tilde{P}(\vX^t)$ and, if the estimator is flawed, \cref{alg:differentiable-temporal-recourse} could provide sub-optimal recourse. 
Our experiments on synthetic and realistic datasets confirm that time presents a non-trivial challenge for AR and show how an estimator approximating $P(\vX^t)$ can still be useful in some scenarios.
However, practical temporal recourse needs reliable time series forecasting, which is well-known to be challenging in various settings \cite{makridakis2020m4}, because of issues like concept drift \cite{gama2014survey}. Additionally, it requires effective implementations of causal inference over time \cite{cinquini2024practical}, which is an ongoing research topic. 
%
%
%
In practice, we could not fully evaluate the effectiveness of \cref{alg:differentiable-temporal-recourse} in real-world situations as this requires temporal datasets for recourse, which are currently not available or hardly obtainable.
For instance, in the case of lending applications, while banks hold temporal data on loan applicants - \eg records of the same individual reapplying for a loan after some time — these datasets are usually proprietary and not publicly accessible \citep{crupi2024counterfactual,castelnovo2020befair}.
}%
Indeed, the scarcity of suitable data is a well-known issue affecting the evaluation of AR approaches at large \cite{karimi2021algorithmic,karimi2020survey,majumdar2024carma}.
Suitable datasets for temporal recourse tasks, in particular, should at a minimum include the features used by decision-makers to train classifiers. For instance, in credit lending, financial institutions may rely on both frequently updated data (\eg credit card spending patterns) and more static attributes (\eg collateral assets).
To capture evolving trends in data distributions, these datasets will likely need to incorporate not only internal institutional records but also external sources such as governmental statistics (\eg census data).
Moreover, certain features may be considered ``fixed'' and thus not require temporal modeling.
In addition, regulatory frameworks could facilitate access to richer datasets; for example, Article 40 of the European DSA\footnote{\url{https://eur-lex.europa.eu/eli/reg/2022/2065/oj}} permits vetted researchers to access data from large online platforms (\eg Facebook, Twitter) for research purposes.
Extending such provisions to other domains could significantly benefit algorithmic recourse research.
However, privacy and ethical concerns remain central, necessitating rigorous data handling and anonymization practices.

\textbf{Causal models, trends and interventions.}
\GDT{
Our formalization assumes that the stochastic process adheres to the TiMiNo framework \cite{peters2013causal}, but it leaves room for exploring more complex Structural Causal Models (SCMs).
Addressing such complexities would necessitate more versatile models for causal inference in time series, such as integrating classical approaches like VAR \cite{zivot2006vector}. 
Additionally, we do not examine trend models for the classifier $h$ or the cost function $C(\cdot)$, even though such trends may coexist with those of $P(\vX^t)$.
} \GDT{
Lastly, our task also aligns with alternative paradigms, such as the \textit{dynamic treatment regime} \cite{murphy2003optimal}, which emphasize constructing a \textit{sequential policy} rather than a single counterfactual to ensure a desired effect at time $t+\tau$. We believe our theoretical results can extend to these settings, and developing their connection with AR represents a potential future direction.
}

\textbf{Further ethical considerations.}
\GDT{
Our work seeks to achieve \textit{algorithmic contestability} \cite{lyons2021conceptualising} through actionable counterfactual explanations, empowering users to challenge decisions made by machine learning models.
We believe recourse can offer tangible benefits to decision subjects.
For instance, in credit lending, recourse actions could help \textit{scarred borrowers} \cite{cowlingHasPreviousLoan2022}, individuals who self-exclude from (re-)applying due to a perceived negative response from banks, by providing tools that enhance their understanding of the decision-making process.
However, AR introduces both technical and ethical challenges \cite{venkatasubramanian2020philosophical}.
For example, counterfactuals must avoid exhibiting \textit{adversarial characteristics} \cite{leemann2024towards} and should aim to improve the user's ground truth classification.\footnote{\GDT{Consider an imaginary university application screening system that evaluates the number of books owned by an applicant's family, based on the assumption that this metric correlates with academic performance \cite{evans2014booksExposure}. While purchasing a crate of 1000 books might artificially improve the applicant's chances of acceptance, it does not enhance their actual academic abilities (unless, of course, they read the books). Such an action would represent an example of an \textit{adversarial recourse suggestion}.}}
Additionally, fairness considerations must be addressed \cite{von2022fairness} before implementing recourse solutions in real-world settings. 
Related research has explored fairness in sequential decision-making within dynamical systems \cite{huAchievingLongTermFairness2022,rateike2024fairpolicies}, and extending these principles to recourse actions represents a critical step for realistic deployment.
Lastly, recourse methods have been shown to pose risks of \textit{private data leakage} \cite{pmlr-v206-pawelczyk23a}, potentially exposing sensitive user information to malicious entities.
To be effective in the wild, AR applications must take into account all these factors to ensure their effectiveness.
}

\section{Conclusions}
\label{sec:conclusion-limitations}
In this paper, we have investigated the impact of time on algorithmic recourse.
Our formalization of temporal causal recourse extends both counterfactual and sub-population causal AR by modelling the world as a (possibly non-stationary) causal stochastic process $P(\vX^t, Y^t)$.
It allows us to theoretically demonstrate how \evidenzia{standard and robust AR approaches are fragile}, as their solutions become invalid in the presence of trends and future uncertainty. We also show that a simple algorithm, leveraging an estimator of the stochastic process fitted on historical data, can deliver more robust solutions.
Our experiments with causal and non-causal approaches support our findings.
With this work, we aim to highlight the negative impact of time on existing AR approaches while demonstrating how these challenges can be at least partially mitigated by leveraging historical data.

\section*{Acknowledgments}

The authors thank Martin Pawelczyk for the helpful feedback and discussions on an initial version of the manuscript.
The work was partially supported by the following projects: MUR PNRR project FAIR - Future AI Research (PE00000013) funded by the NextGenerationEU (AP and BL), Horizon Europe Programme, grant \#101120237-ELIAS (AP and BL) and grant \#101120763-TANGO (ST, AP and BL).
Views and opinions expressed are however those of the author(s) only and do not necessarily reflect those of the European Union or the European Health and Digital Executive Agency (HaDEA).
Neither the European Union nor the granting authority can be held responsible for them.

\bibliography{references}
\bibliographystyle{plainnat}

\newpage
\appendix
\section{Proofs}
\label{app:proofs}

\begin{proof}
    Consider a TiMINo stochastic process $\{\vX_{t}\}_{t\in\bbN}$ with structural equations:
    \begin{equation}
        X_i^t = f(\vPa_i^{t-p}, \ldots,\vPa_i^{t}) + U_i^t,
        \quad
        U_i^t \sim \calN(\mu_X, \sigma_X),
        \quad
        \mu_X, \sigma_X > 0
    \end{equation}
    where $U_i^t$ have positive mean and variance.
    We prove the proposition by contradiction by looking at the value of the time lag $\tau$.
    For the sake of clarity, we state the proof for $p=1$ (the proof for $p>1$ is similar).
    Assume that we observed a sequence of realizations $\vx^{t-p:t}$ and we can compute the counterfactual distribution $P^{do(\vX^{t+\tau} = \vx^{t+\tau} + \vtheta), \vX^{t-p:t}=\vx^{t-p:t}}(\vX^{t+\tau})$ for any $\tau \geq 0$ and $\vtheta \in \bbR^d$.
    If $\tau = 0$, we can immediately recover the value of all exogenous factors via $u_i^t = x_i^t - f(\vPa_i^{t-p}, \ldots,\vPa_i^{t})$, as we know $\vx^{t-p:t}$.
    Let us assume we can also recover the exogenous factors for a time lag $\tau >0$ even if we did not observe future realizations $\{\vx^{t},\ldots,\vx^{t+\tau}\}$.
    Since we did not observe such realizations, the classical abduction step to recover the exogenous factors within the $[t, t+\tau]$ interval is impossible.
    Recently, \citet{bynum2023counterfactuals} introduced \textit{forward-looking counterfactuals} to overcome this challenge. 
    They postulate we can still compute counterfactuals over unseen realizations by \textit{propagating} the latest exogenous factors we were able to abduce \citep[Section 3]{bynum2023counterfactuals}.
    Similary, we assume we can \textit{propagate} the latest exogenous factors we can recover ($\vu^{t}$) into the future, \eg $U^{t+\tau}_i=u^t_i$ for any $\tau >0$ and $i \in [d]$.
    Since we assume we can compute the exact counterfactual distribution, the only setting where $U^{t+\tau}_i=u^t_i$ holds is if $\sigma_X = 0$ and $\mu_X = 0$.
    However, we initially stated that $\mu_X, \sigma_{X} > 0$, so we have a contradiction.
\end{proof}

\subsection{Proof of Proposition \ref{prop:stationary-equivalence-recourse}}

\begin{proof}
Consider a discrete-time stationary time series $\{(\vx, y)^t\}_{t\in \bbN}$ and wlog consider a fixed linear classifier $h(\vx) = \vbeta^{\top}\vx$ that is \textit{injective}, that is, $\vx \neq \vx' \Rightarrow h(\vx) \neq h(\vx')$.
Given a realization $\vx^t$, let us assume $\vtheta^*$ is the optimal intervention for \cref{eqn:temporal-recourse-obj} at time $t$, but not at time $t + \tau$, with $\tau \in \bbN$.
Thus, either our recourse becomes invalid or exceedingly expensive.
We focus on the former since we assume a fixed cost function such as the $\ell_1$-norm of the intervention $C(\vtheta) = \| \vtheta \|$. 
As a consequence, we have $\bbE[h(\hat{\vx}^{t+\tau}) - h(\hat{\vx}^{t})] \neq 0$ where $\hat{\vx}^{t+\tau} \sim P^{do(\vtheta^*)}(\vX^{t+\tau} \mid \vX_{nd(\calI)}^{t+\tau} = \vx_{nd(\calI)}^{t+\tau})$ and $\hat{\vx}^{t} \sim P^{do(\vtheta^*)}(\vX^{t} \mid \vX_{nd(\calI)}^{t} = \vx_{nd(\calI)}^{t})$. 
We denote with $\calI \subseteq [d]$ the intervention set of the successful intervention $\vtheta^*$, and with $do(\vtheta^*) = do(\vX_\calI^t = \vx_{\calI}^t + \vtheta^*)$ the corresponding soft intervention. 
Consider the case in which our intervention does not achieve recourse after $\tau$ time steps.
We can decompose the previous expectation as follows:
\begin{align}
    \bbE[h(\hat{\vx}^{t+\tau}) - h(\hat{\vx}^{t})] &= \bbE[\vbeta^{\top}\hat{\vx}^{t+\tau}-\vbeta^{\top}\hat{\vx}^{t}] \\
    &= \vbeta^{\top}\bbE[\tilde{\vx}^{t+\tau}-\tilde{\vx}^{t}]
\end{align}
Thus, since $\vtheta^*$ is not optimal for $t+\tau$, we must have $\bbE[\tilde{\vx}^{t+\tau}-\tilde{\vx}^{t}] \neq 0$.  This, however, is a contradiction since we assume the time series is stationary.
It follows that, since $\tau$ is arbitrary, for any $t$, the corresponding optimal solution $\vtheta^*$ is also optimal for all $\tau > 0$.
Now, this also means that $\vtheta^*$ is also an optimal solution for the classical recourse optimization problem \cref{eqn:causal-recourse-equation} as long as we optimize \cref{eqn:causal-recourse-equation} by considering $P(\vX, Y) = P(\vX^t, Y^t)$ for any time steps $t \in \bbN$.
\end{proof}

\subsection{Full derivations for Example \ref{example:fragility-of-recourse}}

\begin{proof}
Consider a trend-stationary stochastic process defined by these structural equations:
\[
    \begin{aligned}
        & X^t = \alpha X^{t-1} + m(t) + U_X^t,
            & U_X^t \sim \calN(\mu_X,\sigma_X) \qquad
        \\
        & Y^t = \beta X^t + U_Y^t,
            & U_Y^t \sim \calN(0,1) \qquad
    \end{aligned} 
\]
for all $t \in \bbN$, $\alpha \in (0, 1)$ and $\beta \in \bbR$.
The function $m(t): \bbR \rightarrow \bbR$ represents a \textit{trend} independent of $X^t$ and $Y^t$.
We consider a linear trend $m(t) = -ct + U_m^t$, where $U_m^t \sim \calN(\mu_m,\sigma_m)$ and $c \in \bbR^+$.
Given a realization $x^{t-1}$, the state of $X^{t+\tau}$ admits the closed-form expression:
\begin{align}
    \textstyle
    X^t &= \alpha x^{t-1} + m(t) + U^t_X \\
    X^{t+1} &= \alpha^2 x^{t-1} + \alpha m(t) + \alpha U^t_X + m(t+1) + U^{t+1}_X \\
     \vdots & \\
     X^{t+\tau} &= \alpha^{\tau+1}x^{t-1} + \sum_{i=0}^{\tau} \alpha^{\tau-i} \left(  m(t+i)+ U_X^{t+i} \right)
\end{align}
Hence, the expectation of $Y^{t+\tau}$ with respect to the interventional distribution $P^{do(\theta)}(X^t, Y^t)$ is:
\begin{align*}
    \bbE[Y^{t+\tau}]
        & = \bbE[\beta X^{t+\tau} + U_Y^{t+\tau}] \\
        & = \beta \bbE[X^{t+\tau}] + \bbE[U_Y^{t+\tau}] \\
        &\textstyle \overset{(i)}{=} \beta \bbE[\alpha^{\tau+1}x^{t-1} + \sum_{i=0}^{\tau} \alpha^{\tau-i} \left(  m(t+i)+ U_X^{t+i} \right)]
        \\
        &\textstyle \overset{(i)}{=} \beta \alpha^{\tau+1}x^{t-1} + \sum_{i=0}^{\tau} \alpha^{\tau-i} \left(  \bbE[m(t+i)] + \bbE[U_X^{t+i}] \right) \\
        & \textstyle = \beta \left( \alpha^{\tau+1}x^{t-1} + \sum_{i=0}^{\tau} \alpha^{\tau-i} (-c(t+i) + \mu_m + \mu_X) \right)
\end{align*}
Here, $(i)$ follows by construction since $\bbE[U_Y^{t+\tau}]=0$ for all $t$, $\tau$ and $i \in [d]$.
We now consider the following fixed classifier $\sigma(Y^t \mid X^t)$ where $\sigma(x) = 1/(1+e^{-x})$. Thus, the expectation over the classifier output becomes:
\[
    \textstyle
    \bbE[h(X^{t+\tau})]
        = \sigma \left( \beta \left( \alpha^{\tau+1}x^{t-1} + \sum_{i=0}^{\tau} \alpha^{\tau-i} (-c(t+i) + \mu_m + \mu_X) \right) \right)
\]
Given that we are considering soft interventions, we consider the cost function $C(\hat{x}^{t+\tau}, x^{t+\tau}) = \hat{x}^{t+\tau} - x^{t+\tau} = \theta$ since $\hat{x}^{t+\tau} = x^{t+\tau} + \theta$.  
Given that $\sigma(x) \geq 1/2$ if and only if $x \geq 0$, we have that the optimal intervention $\theta^{t+\tau} \in \bbR$ for which we have $\bbE[h(X^{t+\tau} + \theta)] \geq 1/2$ can be expressed as:
\[
    \textstyle
    \theta^{t+\tau} = -\alpha^{\tau+1}x^{t-1} -\sum_{i=0}^{\tau} \alpha^{\tau-i} (-c(t+i) + \mu_m + \mu_X)
\]
\end{proof}

\subsection{Proof of Proposition \ref{prop:robust-recourse-fails-over-time}}

We can prove \cref{prop:robust-recourse-fails-over-time} by showing how we can \textit{always} find a simple trend invalidating \textit{any} (robust) intervention.
\begin{proof}
Let us consider a trend-stationary stochastic process $P(\vX^t, Y^t)$ and fixed injective classifier $h$ approximating $P(\vX^t \mid Y^t)$.
We denote with $\vm(t): \bbN^d \rightarrow \bbR^d$ the $d$-variate trend function where $m_i(t)$ is the trend component for a single random variable $X_i^t$ for any $i \in [d]$ and $t \in \bbN$. 
Given a negatively classified instance $\vx^t$, assume $\vtheta$ is the optimal robust intervention for a fixed $\epsilon > 0$ and for the timestep $t$.

Consider the following trend function $\vm(t) = \bm{1}\{t \geq \tau\}(-\vtheta)$ which is adding the inverse of the optimal intervention if an only if $t \geq \tau$.
Specifically, we define each trend component as $m_i(t) = \bm{1}\{t \geq \tau\}(-\theta_i)$.
If our stochastic process exhibits such a trend, then, for any fixed $\tau > 0$, the robust intervention is invalid \eg $\bbE[h(\hat{\vx}^{t+\tau}])] < 1/2$.
Moreover, we can always build such a trend for any $\vtheta$ and any unrestricted trend-stationary stochastic process. 

\end{proof}

\subsection{Proof of Theorem \ref{theorem:invalidation-rate}}
\label{app:proof-invalidation-rate}

\begin{proof}
We first apply the following substitutions (a) $\vbeta' = \vbeta^{t+\tau}$ and $\vx' = \hat{\vx}^{t+\tau}$ (b) $\vbeta = \vbeta^{t}$ and $\vx = \hat{\vx}^{t}$, to improve the clarity of the proof.
Further, consider $\Delta h = \bbE[|h^{t+\tau}(\hat{\vx}^{t + \tau})-h^{t}(\hat{\vx}^t)|]$.
Then, the proof is the following:
    \begin{equation*}
        \begin{aligned}
            \Delta h &= \bbE[ \left| \dprod{\vbeta'}{\vx'} - \dprod{\vbeta}{\vx} \right|]\\
            & = \bbE[ | \dprod{\vbeta'}{\vx'} \\
            & \qquad + \dprod{\vbeta}{\vx'} - \dprod{\vbeta}{\vx'} - \dprod{\vbeta}{\vx} + \dprod{\vbeta}{\vx'} - \dprod{\vbeta}{\vx'}  | ] \nonumber \\
            & = \bbE[ \left| \dprod{\vbeta'-\vbeta}{\vx'}+\cancel{\dprod{\vbeta}{\vx'}} + \dprod{\vbeta}{\vx'-\vx} - \cancel{\dprod{\vbeta}{\vx'}} \right|] \\
            & \leq \bbE[ \left| \dprod{\vbeta'-\vbeta}{\vx'}\right|] + \bbE[\left| \dprod{\vbeta}{\vx'-\vx}  \right|] \\
            & \overset{(i)}{\leq} \bbE[ \norm{\vbeta'-\vbeta} \cdot \norm{\vx'}] + \bbE[\norm{\vbeta}\cdot \norm{\vx'-\vx}] \\
            & \overset{(ii)}{\leq} k\sqrt{d}\cdot \bbE[ \norm{\vbeta'-\vbeta}] + k\sqrt{d}\cdot\bbE[ \norm{\vx'-\vx}] \\
            & = k\sqrt{d} \cdot \left( \expect{\norm{\vbeta^{t+\tau}-\vbeta^t}} + \expect{\norm{\hat{\vx}^{t+\tau}-\hat{\vx}^t}} \right)
        \end{aligned}
    \end{equation*}
where (\textit{i}) follows from the Cauchy-Schwarz inequality and (\textit{ii}) from the bounds we placed on $\vX$ and $\vbeta$ (\eg since $-\vk \leq \vbeta \leq \vk$ and since $|\vbeta| = d$ we have $\max_\vbeta \norm{\vbeta} = k\sqrt{d}$).
Moreover, since $h^t$ is trained over a fixed dataset $\calD^t$, we have that $\vbeta^t \bot\; \vX^t \mid \calD^t $ for all $ t \in \bbN$. We stress that the bounds placed on $\vX$ and $\vbeta$ enable us to constrain the $\Delta h(\vtheta; \tau)$ variation. In the unbounded case, where $k \to \infty$, clearly no upper bound is possible.
\end{proof}

\subsection{Proof of Corollary \ref{corollary:invalidation-rate-trend}}

We begin the Proof of Corollary \ref{corollary:invalidation-rate-trend} by starting from the previous proof for \cref{theorem:invalidation-rate} (given in \cref{app:proof-invalidation-rate}).
Please recall that a stochastic process $P(\vX^t)$ is trend-stationary when it can be expressed as $\vX^t = \vm(t) + \ve^t$, where $\vm(t)$ is a (non-)linear trend function and $\ve^t$ is a stationary stochastic process. 
In the following, we denote with $\tilde{\vx}$ the stationary part of the stochastic process, and we consider \textit{deterministic} trend functions. 
\begin{proof}
Let us assume that each random variable $X^t_i$ can be described as a trend-stationary univariate stochastic process.
Thus, let us define with $\vm(t) = \{m_i(t)\}^d_{i=1}$ the trend function, where $m_i(t)$ the trend component for the $i$-th variable. 
Then, we define as $m^* = m_j$ the trend with the largest difference $j = \argmax_{i \in [d]} m_i(t+\tau) - m_i(t)$ for $t \in \bbN$ and $\tau > 0$.
Further, consider $\Delta h = \bbE[|h^{t+\tau}(\hat{\vx}^{t + \tau})-h^{t}(\hat{\vx}^t)|] $.
The derivation for the upper bound is:
        \begin{align*}
            \Delta h &\leq k\sqrt{d}  \left( \expect{\norm{\vbeta^{t+\tau}-\vbeta^t}} + \expect{\norm{\hat{\vx}^{t+\tau}-\hat{\vx}^t}} \right) \quad \hbox{(\cref{theorem:invalidation-rate})} \\
            &\overset{(i)}{=} k\sqrt{d}  \left( \expect{\norm{\vbeta^{t+\tau}-\vbeta^t}} + \expect{\norm{\tilde{\vx}^{t+\tau} + \vm(t+\tau) -\tilde{\vx}^t - \vm(t)}} \right) \\
            &= k\sqrt{d}  \left( \expect{\norm{\vbeta^{t+\tau}-\vbeta^t}} + \expect{\norm{\tilde{\vx}^{t+\tau}-\tilde{\vx}^t}} + \expect{\norm{\vm(t+\tau)- \vm(t)}}  \right) \\
            &\overset{(ii)}{=} k\sqrt{d}  \left( \expect{\norm{\vbeta^{t+\tau}-\vbeta^t}} + \expect{\norm{\vm(t+\tau)- \vm(t)}}  \right) \\
            &= k\sqrt{d}  \left( \expect{\norm{\vbeta^{t+\tau}-\vbeta^t}} + \sqrt{\sum_{i=1}^d (m_i(t+\tau)- m_i(t))^2} \right) \\
            &\overset{(iii)}{\leq} k\sqrt{d}  \left( \expect{\norm{\vbeta^{t+\tau}-\vbeta^t}} + \sqrt{\sum_{i=1}^d (m^*(t+\tau)- m^*(t))^2} \right) \\
            &= k\sqrt{d}  \left( \expect{\norm{\vbeta^{t+\tau}-\vbeta^t}} + \sqrt{d(m^*(t+\tau)- m^*(t))^2} \right) \\
            &= k\sqrt{d}  \left( \expect{\norm{\vbeta^{t+\tau}-\vbeta^t}} + \sqrt{d}(m^*(t+\tau)- m^*(t))  \right) \\
            &= k\sqrt{d}  \expect{\norm{\vbeta^{t+\tau}-\vbeta^t}} + kd\left(m^*(t+\tau)- m^*(t)\right) 
        \end{align*}
where $(i)$ and $(ii)$ follows from the definition of a \textit{trend-stationary} stochastic process.
Then, $(iii)$ follows since we can substitute each univariate trend $m_i(t)$ with the maximum $m^*(t)$ for the time step $t$ within the $\ell_2$-norm. 
\end{proof}

\section{Implementation Details for the Experiments With Synthetic and Real Data}
\label{app:implementation_details_experiments}

In this section, we describe the synthetic and realistic stochastic processes we used in our experiments (cf. \cref{sec:experiments}). We also describe the training steps and generative model used to approximate the SCMs. Lastly, we report several technical implementation details.  

\subsection{Synthetic Additive Trend Function}
We want our causal stochastic process to describe an environment where the changes to obtain a positive classification fluctuate over time.
For example, as an individual ages, it will become less likely for her to repay her loan in the case of a loan application (based on the survival rate of the population).
We define an additive trend function $m(t): \bbN \rightarrow \bbR^+ $, which is a linear combination between a \textit{linear} and \textit{seasonal} trend. 
We control the mixture of each component with two parameters, $\beta_l \in \bbR^+$ and $\beta_s \in \bbR^+$, respectively.
We also consider an additional parameter $\alpha \in [0,1]$ controlling the trend's strength over the stationary causal process.
\begin{equation}
    m(t) = \alpha \cdot \left( \beta_{l} \cdot \min(0.05\cdot t, 10) + \beta_{s} \cdot | \sin(0.5 \cdot t) |    \right)
\end{equation}
In our experiments, we set $\beta_l \in \{0, 1\}$ and $\beta_s \in \{0, 1.5\}$ for the linear ANM, and $\beta_l \in \{0, 2\}$ and $\beta_s \in \{0, 5\}$ for the non-linear ANM.
For the realistic experiments, we set $\beta_l, \beta_s \in \{0, 1\}$ for \texttt{Adult}, $\beta_l \in \{0, 0.3\}$ and $\beta_s \in \{0, 1\}$ for \texttt{COMPAS}, and $\beta_l \in \{0, 0.5\}$ and $\beta_s \in \{0, 5\}$ for \texttt{Loan}.

\subsection{Causal Graphs for the Experiments}

For the synthetic experiments, we considered the synthetic 3-variables additive noise models (ANMs) from \cite{karimi2020algorithmic}.
We extended them by transforming them into trend-stationary stochastic processes, by adding an autoregressive component and the trend $m(t)$ on the 3rd feature.
If $\alpha=0$, both ANMs give rise to stationary time series. 

\textbf{Linear Additive Noise Model.} 
\begin{equation}
\begin{aligned}
    & X^t_1 = 0.5\cdot X^{t-1}_1 + U^t_1\\
    & X^t_2 = 0.5\cdot X^{t-1}_2 - 0.25\cdot X^{t}_1 + U^t_2  \\
    & X^t_3 = 0.5\cdot X^{t-1}_3 + 0.05\cdot X^{t}_1 + 0.25\cdot X^{t}_1 - m(t) + U^t_3
\end{aligned}
\end{equation}
where $U^t_1 \sim \mathrm{MoG}(\calN(-1, 1.5), \calN(1,1))$, $U_2^t \sim \calN(0, 0.1)$ and $U^t_3 \sim \calN(0, 1)$.

\textbf{Non-linear Additive Noise Model.}
\begin{equation}
\begin{aligned}
    & X^t_1 = 0.5\cdot X^{t-1}_1 + U^t_1 \\
    & X^t_2 = 0.5\cdot X^{t-1}_2 -1 \frac{3}{1+ e^{-2X^{t}_1}} + U^t_2 \\
    & X^t_3 = 0.5\cdot X^{t-1}_3 + 0.05\cdot X^{t}_1 + 0.25\cdot (X^{t}_1)^2 - m(t) + U^t_3
\end{aligned}
\end{equation}
where $U^t_1 \sim \mathrm{MoG}(\calN(-2, 1.5), \calN(1,1))$, $ U_2^t \sim \calN(0, 0.1)$ and $U^t_3 \sim \calN(0, 1)$.

\textbf{Label function.} 
We also consider the following conditional distribution $P(Y^t \mid \vX^t)$, again taken from \cite{karimi2020algorithmic}, which produces roughly balanced groups:
\begin{equation}
    Y^t \sim \mathrm{Binomial}\left( 1/ \left(1+ \mathrm{exp}(-2.5 \cdot (X^t_1 + X^t_2 + X^t_3)/\rho)\right) \right)
\end{equation}
where $\rho$ is the empirical mean of $X^t_1 + X^t_2 + X^t_3$ for $t=0$. 
We use the label function only to train the classifier $h$ at time $t=0$, then it is discarded and we rely only on $h$ for our experiments. 

\subsection{Causal Graphs for the Realistic Experiments}

We now describe the design choices for the realistic datasets. 
We tried to find a balance between realism and simplicity, to provide scenarios close to potential real-world situations, that, however, we can easily control for our experiments. 
Therefore, some of the design choices might not represent faithfully how the system can evolve in real life. 

\textbf{\texttt{Adult} \citep{Dua:2019}}. We use the features and causal graph defined by \cite{nabi2018fair}. 
\cref{app:causal_graph_adult} shows the full causal graph.
We have the following features: $S$ (sex), $A$ (age), $US$ (resident of the United States of America), $M$ (married), $E$ (education level) and $H$ (working hours per week). 
The features $S$, $US$, and $M$ are categorical variables, the others are represented as continuous random variables.
We assume $S$, $A$, $US$ and $M$ remain fixed over time.
The only actionable features are the education level $E$, and working hours per week $H$.
We apply a decreasing trend to $H$.
We employ non-linear structural equations $f_M$, $f_E$ and $f_H$ by using pre-trained 3-layer MLPs, trained on the original dataset, from \cite{dominguez2022adversarial}. 
Moreover, as a label function, we also employ a classifier $h$ taken by \cite{dominguez2022adversarial} to label the examples at $t=0$.

\begin{equation}
    \begin{aligned}
        & S^t = \bm{1}\{t>0\} \cdot S^{t-1} + \bm{1}\{t=0\} \cdot U_S  \\
        & A^t = \bm{1}\{t>0\} \cdot U^{t-1} + \bm{1}\{t=0\} \cdot U^t_A  \\
        & US^t = \bm{1}\{t>0\} \cdot US^{t-1} + \bm{1}\{t=0\} \cdot U_{US}  \\
        &M^t = \bm{1} \{ \sigma( f_M(S^t, A^t, US^t) ) > 1/2 \} & \\
        &E^t = 0.5\cdot E^{t-1} + f_E(S^t, A^t, US^t, M^t) + U_E & \\
        &H^t = 0.5\cdot H^{t-1} + f_H(S^t, A^t, US^t, M^t) -m(t) + U_H
    \end{aligned}
    \label{app:causal_graph_adult}
\end{equation}
where the exogenous factors are $U_S \sim \mathrm{Binomial}(0.9)$, $U^t_A \sim \calN(0,1)$, $U_{US} \sim \mathrm{Binomial}(0.9)$, $U_E \sim \calN(0,1)$ and $U_H \sim \calN(0,1)$.  

\textbf{\texttt{COMPAS} \citep{angwin2016machine}}. We use the features and causal graph defined by \cite{nabi2018fair}. 
\cref{app:causal_graph_compas} shows the full causal graph.
We have the following features: $S$ (sex), $A$ (age), $C$ (ethnicity, if caucasian or not), and $P$ (prior counts). 
The feature $S$ is categorical, and the others are represented as continuous random variables.
We assume $S$, $A$, and $C$ remain fixed over time, and that the only actionable feature is the prior count $P$.
We apply an increasing trend to $P$.
As we did for \texttt{Adult}, we obtain the pre-trained non-linear structural equations $f_C$ and $f_P$ and label function from \cite{dominguez2022adversarial}.

\begin{equation}
    \begin{aligned}
        & S^t = \bm{1}\{t>0\} \cdot S^{t-1} + \bm{1}\{t=0\} \cdot U_S \\
        & A^t = \bm{1}\{t>0\} \cdot U^{t-1} + \bm{1}\{t=0\} \cdot U^t_A \\
        & C^t = \bm{1}\{t>0\} \cdot C^{t-1} + \bm{1}\{t=0\} \cdot \left(\sigma(f_{C}(S^t, A^t)) + U^t_C \right) \\
        &P^t = 0.5\cdot P^{t-1} + f_P(S^t, A^t, C^t) + m(t) + U^t_P\\
    \end{aligned}
    \label{app:causal_graph_compas}
\end{equation}
where the exogenous factors are $U_S \sim \mathrm{Binomial}(0.8)$, $U^t_A \sim \mathrm{Poisson}\allowbreak(1)$,  $U^t_{C} \sim \calN(0, 1)$, and $U^t_P \sim \calN(0,1)$

\textbf{\texttt{Loan} \citep{karimi2020algorithmic}}. We use the causal graph defined by \cite{karimi2020algorithmic} presenting a semi-synthetic loan approval scenario inspired by the German Credit dataset \citep{statlog_(german_credit_data)_144}.
For the structural equations, we use the one adapted by \cite{dominguez2022adversarial} in their paper. 
\cref{app:causal_graph_loan} shows the full causal graph. 
We have the following features: $G$ (gender), $A$ (age), $E$ (education level), $L$ (loan amount), $D$ (loan duration), $I$ (income) and $S$ (savings). 
$G$ is a categorical variable, while the rest are considered continuous. Moreover, $G$ remains fixed over time. 
We assume the only actionable features are $S$ and $I$.
We apply a trend to the income $I$.

\begin{equation}
    \begin{aligned}
    G^t &= \bm{1}\{t>0\} \cdot G^{t-1} + \bm{1}\{t=0\} \cdot U^t_G \\
    A^t &= 0.5\cdot A^{t-1} -35+U^t_A\\
    E^t &= 0.5 E^{t-1} -0.5 + \left(1 + e^{-\left(-1 + 0.5 G^t + \left(1 + e^{- 0.1 A^t}\right)^{-1} + U_E\right)}\right)^{-1}\\
    L^t &= 0.5\cdot L^{t-1} + 1 + 0.01 (A^t - 5) (5 - A^t) + G^t + U_L \\
    D^t &= 0.5\cdot D^{t-1} -1 + 0.1A^t + 2G^t + L^t + U_D \\
    I^t &= 0.5\cdot I^{t-1} -4 + 0.1(A^t + 35) + 2G^t + G^t E^t + U_I - m(t) \\
    S^t &= 0.5\cdot S^{t-1}  -4 + 1.5 \cdot \bm{1}{\{I^t > 0\}} \cdot I^t + U_S 
    \end{aligned}
    \label{app:causal_graph_loan}
\end{equation}
where the exogenous factors are $U^t_G \sim \mathrm{Binomial}(0.5)$, $U^t_A \allowbreak \sim \allowbreak  \text{Gamma} \allowbreak (10, \allowbreak 3.5)$, $U_E\sim \calN(0, \sqrt{0.25})$, $U_L\sim \calN(0, 2)$, $U_D\sim \calN(0, 3)$, $U_I\sim \calN(0, 2)$, and $U_S\sim \calN(0, 5)$. 
We sample the labels from the function defined by \cite{karimi2020algorithmic}:
\begin{equation}
    Y^t\sim \text{Bernoulli}\left(\left(1+e^{-0.3(-L^t-D^t+I^t+S^t+IS^t)}\right)^{-1}\right)
\end{equation}

\subsection{On Learning an Approximate SCM}
\label{app:sec:generative_model}

We now describe the simple generative model we used in the experiment in \cref{sec:real-experiment}.
Similarly to \cite{karimi2020algorithmic} and \cite{dominguez2022adversarial}, we \textit{approximate} the structural equations in a data-driven manner.
We assume we can represent the actionable features $X_i$ as Gaussian random variables $\calN(\mu^t_i, 1)$ with constant variance and time-dependent $\mu^t$. 
For each random variable, we define the mean as the output of a regressor $f_i$ taking as input: the autoregressive component $\vX_i^{t-1}$, the parents $\vPa^t_i$ and the time $t$.
Thus, we obtain the following structural equations:
\begin{equation}
    \begin{aligned}
        X_i^t = \mu_i^t + U^t_i \qquad U^t_i \sim \calN(0, 1) \quad \mu^t_i = f_i(X_i^{t-1}, \vPa_i^t, t)
    \end{aligned}
\end{equation}
Similarly to a conditional VAE \citep{sohn2015learning}, we can both sample new instances from the approximate SCMs, but we can also compute the interventional or counterfactual distributions.

\subsection{Technical Details and Code}

We based our implementation of \CAR, \SAR, \IMF and \TSAR following \textit{adversarial robust algorithmic recourse} \citep{dominguez2022adversarial}. 
Namely, we leveraged their implementation\footnote{\url{https://github.com/RicardoDominguez/AdversariallyRobustRecourse}} and we adapted their code to work with time-based uncertainty sets (cf. \cref{sec:method}).
Moreover, we also used their causal graph implementations, pre-trained models, and preprocessing steps as a starting point for building our stochastic processes.
In the case of the synthetic datasets, we instead looked at \cite{karimi2020algorithmic} original implementation\footnote{\url{https://github.com/charmlab/recourse}}.
Our code and experimental results will be released on \texttt{Github} with a permissive license\footnote{\url{https://github.com/unitn-sml/temporal-algorithmic-recourse}}.
Lastly, we run our experiments on a Linux machine (Ubuntu 22.04, 4 LTS) with 32 CPU cores and 125 GB of RAM.
Our implementation is written in Python 3.10, using standard scientific and deep learning libraries such as \texttt{numpy} \citep{harris2020array} and \texttt{PyTorch} \citep{paszke2019pytorch}. 
The various hyperparameters are duly specified in the source code, but we report the most important here for clarity.

\textbf{\cref{alg:differentiable-temporal-recourse} hyperparameters.} We approximate $B(\vx^t; \tau)$ by sampling only 20 instances for all datasets and we set the number of \textit{epochs} to $N=30$. 
As a penalty, we set $\lambda=1$ for the Lagrangian (line 5, \cref{alg:differentiable-temporal-recourse}).
Please notice that \cite{dominguez2022adversarial} uses a \textit{decaying rate} to reduce the impact of the cost function on the loss $\calL$ after each epoch (we kept the original hardcoded value of $0.02$). 
As learning rate, we set $\eta=0.5$ for the synthetic experiments, and $\eta=3$ for the realistic datasets. The learning rate is the same for all the methods. We did not perform a full grid search over the parameter space, since we found empirically our chosen hyperparameters were giving satisfactory performances. 

\textbf{Classifiers $h$.} For each setting, we trained a 3-layered MLP approximating $P(Y^t \mid \vx^t)$, via \textit{empirical risk minimization} by sampling a given dataset for $t=0$. We use stochastic gradient descent (SGD) to minimize the binary cross entropy loss(\eg \texttt{torch.nn.BCELoss}\footnote{\url{https://pytorch.org/docs/stable/generated/torch.nn.BCELoss.html}})
\begin{equation}
    \calL = -\frac{1}{N}\sum_{\vx^0, y^0 \in \; \hbox{batch}}(y^0\log{h(\vx^0)} + (1-y^0)\log{(1-h(\vx^0))}
\end{equation}
 where $y^0$ is the ground truth label. 
In our experiments, we set the batch size to $100$, the number of epochs to 15 and the learning rate to $0.001$, for all datasets.
The accuracy of the models for a single seed are: $0.847$ (Linear ANM), $0.963$ (Non-Linear ANM), $0.817$ (\texttt{Adult}), $0.645$ (\texttt{COMPAS}) and $0.842$ (\texttt{Loan}). 

\textbf{Approximate structural equations.}
In our experiments, we consider only linear $f_i$. We train each $f_i$ via \textit{empirical risk minimization} following the procedure outlined in \cref{sec:real-experiment}. For each feature $i \in [d]$, and for each epoch, we consider a batch $\{ (x_i^t, x_i^{t-1}, \vPa_i^t, \allowbreak t)_j \allowbreak \}_{j=1}^b$ and we minimize the \textit{mean squared error} between the ground truth $x_i^t$ and the model output $f_i(X_i^{t-1}, \vPa_i^t, t)$ with stochastic gradient descent.
We fix the batch size to 100, the learning rate to $0.001$ and the number of epochs to 15 for all settings. 
\GDT{
We report here the \textit{mean squared error} over 50 timesteps (2000 individuals) for the approximate SCMs we used in \cref{sec:real-experiment}. We compute the MSE for each feature for each timestep, and then we average. The empirical average MSE and standard deviation over 10 runs is: $1.162 \pm 0.005$ (\texttt{Adult}), $258.226 \pm 7.328$ (\texttt{COMPAS}) and $10.447 \pm 0.037$ (\texttt{Loan}).
}

\begin{figure*}[h]
    \centering
    \includegraphics[width=0.8\linewidth]{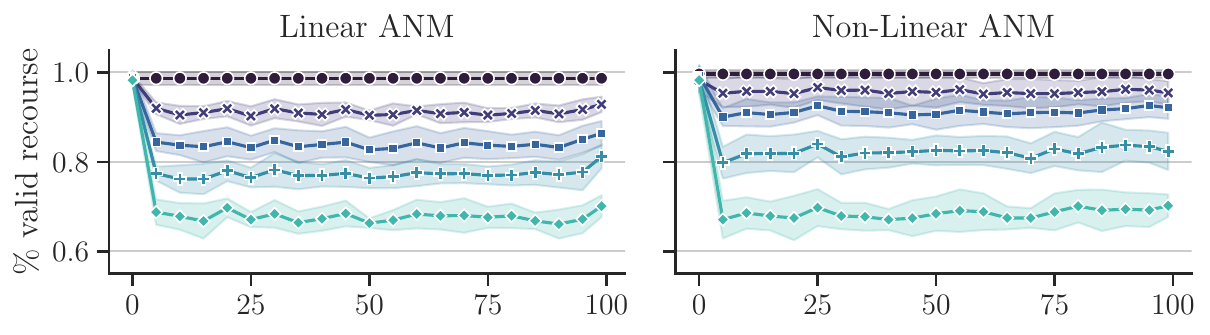}
    \caption{\textbf{Effect of uncertainty on counterfactual AR over time.}
    Empirical average validity and standard deviation over 10 runs of \textit{robust} counterfactual algorithmic recourse (\CAR) for $t \in \{0, 100\}$. We vary the variance $\sigma_U$ of the exogenous factors of the stochastic process.
    Legend ($\sigma_U$): 
    \raisebox{0.5ex}{\fcolorbox[HTML]{FFFFFF}{2E1E3B}{\rule{0pt}{1pt}\rule{1pt}{0pt}}} $0$
    \raisebox{0.5ex}{\fcolorbox[HTML]{FFFFFF}{413D7B}{\rule{0pt}{1pt}\rule{1pt}{0pt}}} $0.3$
    \raisebox{0.5ex}{\fcolorbox[HTML]{FFFFFF}{37659E}{\rule{0pt}{1pt}\rule{1pt}{0pt}}} $0.5$
    \raisebox{0.5ex}{\fcolorbox[HTML]{FFFFFF}{348FA7}{\rule{0pt}{1pt}\rule{1pt}{0pt}}} $0.7$
    \raisebox{0.5ex}{\fcolorbox[HTML]{FFFFFF}{40B7AD}{\rule{0pt}{1pt}\rule{1pt}{0pt}}} $1.0$.
    }
    \label{app:fig:uncertainty-counterfactual-recourse-extended}
\end{figure*}

\begin{figure*}[h]
\centering
\begin{minipage}{0.65\textwidth}
\centering
\includegraphics[width=\linewidth]{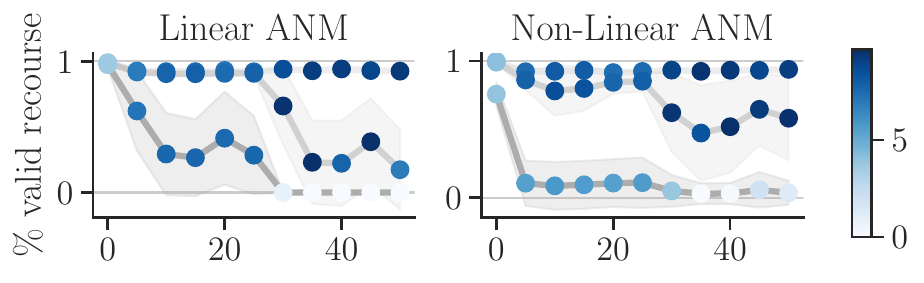}
\end{minipage}
\begin{minipage}{0.95\textwidth}
\centering
\includegraphics[width=\linewidth]{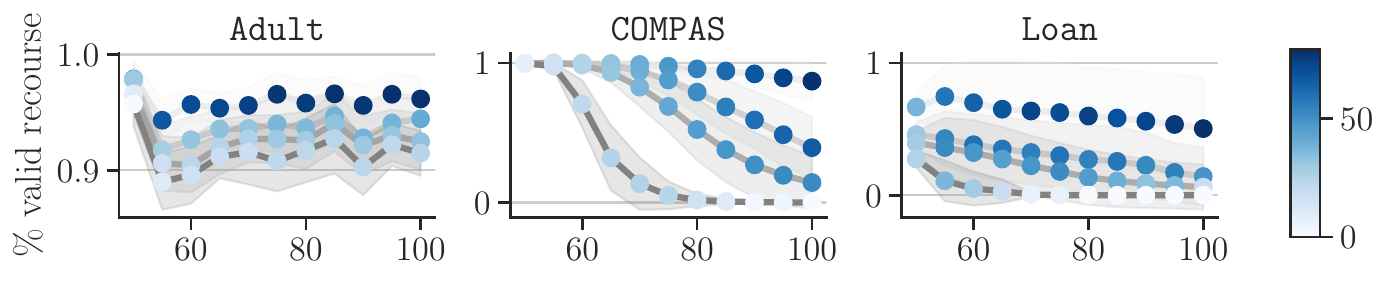}
\end{minipage}

\caption{
\textbf{Trade-off between cost and validity for realistic and synthetic datasets.}
We report the empirical average validity for \cref{alg:differentiable-temporal-recourse} under the linear+seasonal trend ($\alpha=1.0$) when varying the $\lambda$ for synthetic (top) and realistic (bottom) time series.
We consider a non-linear classifier $h$ (3-layer neural network) for each setting. 
Each dot represents the empirical average cost of the interventions achieving recourse. A darker dot implies a larger cost. 
We represent the validity for each $\lambda$ as a grey line and the standard deviation over 10 runs as a shaded area.
Legend ($\lambda$):
    \raisebox{0.5ex}{\fcolorbox[HTML]{FFFFFF}{EDEDED}{\rule{0pt}{1pt}\rule{5pt}{0pt}}} 1
    \raisebox{0.5ex}{\fcolorbox[HTML]{FFFFFF}{D1D1D1}{\rule{0pt}{1pt}\rule{5pt}{0pt}}} 3
    \raisebox{0.5ex}{\fcolorbox[HTML]{FFFFFF}{ADADAD}{\rule{0pt}{1pt}\rule{5pt}{0pt}}} 5 (Synthetic time series),
    \raisebox{0.5ex}{\fcolorbox[HTML]{FFFFFF}{EDEDED}{\rule{0pt}{1pt}\rule{5pt}{0pt}}} 0.01
    \raisebox{0.5ex}{\fcolorbox[HTML]{FFFFFF}{D1D1D1}{\rule{0pt}{1pt}\rule{5pt}{0pt}}} 0.05
    \raisebox{0.5ex}{\fcolorbox[HTML]{FFFFFF}{ADADAD}{\rule{0pt}{1pt}\rule{5pt}{0pt}}} 0.1
    \raisebox{0.5ex}{\fcolorbox[HTML]{FFFFFF}{828282}{\rule{0pt}{1pt}\rule{5pt}{0pt}}} 0.5 (Realistic time series).
}
\label{app:fig:tradeoff-cost-validity}
\end{figure*}

\section{\GDT{Additional Empirical Results on the Effect of Uncertainty on Counterfactual Recourse Over Time}}
\label{app:extended-empirical-results-car}

We present the extended results of the experiment measuring empirically the impact of the uncertainty in \CAR (\cref{sec:counterfactual-tar}).
The experimental setting and evaluation procedure are the same as \cref{sec:synthetic-experiment}.
\cref{app:fig:uncertainty-counterfactual-recourse-extended} shows the empirical average validity of \CAR's recourse over time $t \in \{0, 100\}$.
The plot shows how uncertainty heavily impacts the recourse validity from the initial time steps as $\sigma_\vU$ grows, even when $P(\vX^t)$ is stationary.

\section{\GDT{Analysis of the Trade-off Between Validity and Cost}}
\label{app:sec:trade-off-validity-cost}

\GDT{
We present an analysis of the trade-off between the validity and cost of recourse generated by \cref{alg:differentiable-temporal-recourse}.
Using the same experimental setup and evaluation procedure detailed in \cref{sec:synthetic-experiment,sec:real-experiment}, we evaluate the impact of varying the $\lambda$ parameter, which governs the strength of the cost constraint (line 5, \cref{alg:differentiable-temporal-recourse}).
For this analysis, we focus on time series with a complex linear and seasonal trend.
The results, shown in \cref{app:fig:tradeoff-cost-validity}, demonstrate a clear cost-validity trade-off across all experimental scenarios: \textbf{cheaper interventions result in lower validity over time}.
As the weight of the cost constraint ($\lambda$) increases during optimization, the interventions become more affordable but exhibit reduced validity.
These findings align with prior research on cost-validity trade-offs in non-temporal settings \citep{pawelczyk2022trade}, extending the observations to temporal scenarios.
}
For example, in \texttt{Adult}, we observe a reduction in the validity over time ($\sim 0.05$), but a decreased cost as shown by the lighter dots.
In \texttt{COMPAS}, the trade-off presents a smoother behaviour since we have only one actionable feature. 
Lastly, we observe how the reduction in validity is not consistent across time series: we suspect this might depend on the quality of the estimator $\tilde{P}(\vX^t)$ and on the decision boundary of the classifier $h$.

\begin{figure*}[h]
    \centering
    \begin{minipage}{\textwidth}
        \centering
        \includegraphics[width=\linewidth]{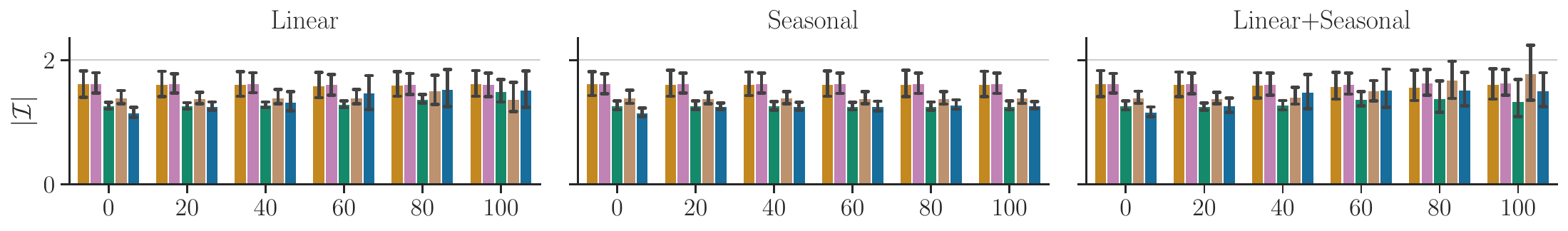}
    \end{minipage}
    \begin{minipage}{\textwidth}
        \centering
        \includegraphics[width=\linewidth]{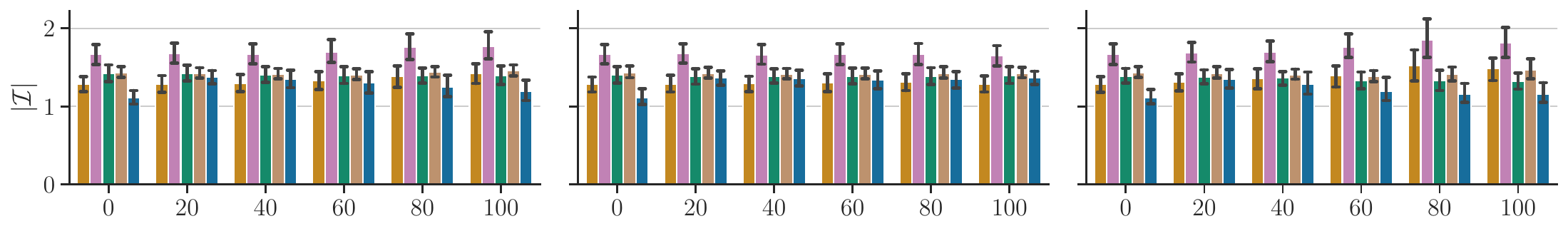}
    \end{minipage}
    \caption{Empirical average sparsity and standard deviation of interventions achieving recourse for all causal recourse methods in the Linear (top) and Non-Linear (bottom) ANMs.
    We report the results for all the available trends $m(t) \in \{ \hbox{Linear}, \hbox{Seasonal}, \hbox{Linear+Seasonal} \}$ and for some time steps $t$.
    Legend:
    \raisebox{0.5ex}{\fcolorbox[HTML]{FFFFFF}{0173B2}{\rule{0pt}{1pt}\rule{5pt}{0pt}}} \TSAR
    \raisebox{0.5ex}{\fcolorbox[HTML]{FFFFFF}{DE8F05}{\rule{0pt}{1pt}\rule{5pt}{0pt}}} \CAR ($\epsilon = 3$)
    \raisebox{0.5ex}{\fcolorbox[HTML]{FFFFFF}{029E73}{\rule{0pt}{1pt}\rule{5pt}{0pt}}} \SAR ($\epsilon = 3$)
     and 
    \raisebox{0.5ex}{\fcolorbox[HTML]{FFFFFF}{CC78BC}{\rule{0pt}{1pt}\rule{5pt}{0pt}}} \CAR ($\epsilon = 5$)
    \raisebox{0.5ex}{\fcolorbox[HTML]{FFFFFF}{CA9161}{\rule{0pt}{1pt}\rule{5pt}{0pt}}} \SAR ($\epsilon = 5$).
    }
    \label{app:fig:sparsity_synthetic}
\end{figure*}

\section{Further Analysis on the Empirical Cost and Sparsity}
\label{app:futher_analysis_cost_sparsity}

In this section, we report further analysis and results when considering the empirical average \textit{cost} and \textit{sparsity} (\# of $\calI$ achieving recourse) of the valid interventions.
They are both common metrics in the algorithmic recourse literature \citep{karimi2020survey, verma2020counterfactual}.

\begin{figure*}[h]
    \centering
    \includegraphics[width=\linewidth]{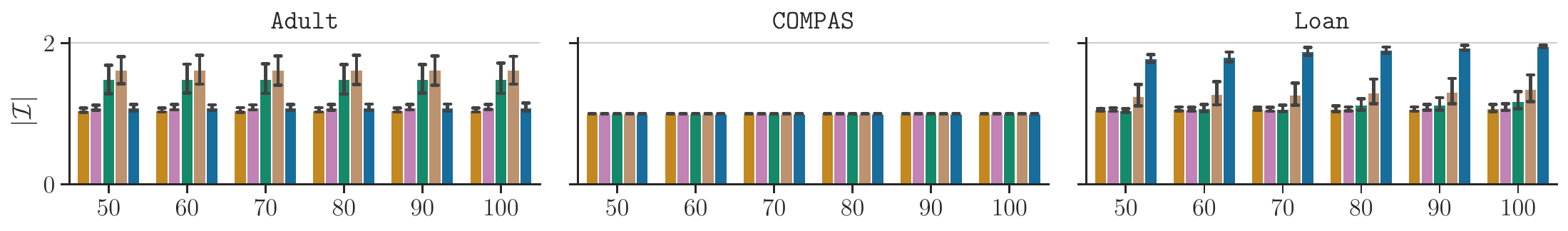}
    \caption{Empirical average sparsity and standard deviation of interventions achieving recourse for all causal recourse methods in the realistic datasets.
    We report the results for $m(t) = \hbox{Linear+Seasonal} $ and for some time steps $t$.
    Legend:
    \raisebox{0.5ex}{\fcolorbox[HTML]{FFFFFF}{0173B2}{\rule{0pt}{1pt}\rule{5pt}{0pt}}} \TSAR
    \raisebox{0.5ex}{\fcolorbox[HTML]{FFFFFF}{DE8F05}{\rule{0pt}{1pt}\rule{5pt}{0pt}}} \CAR ($\epsilon = 3$)
    \raisebox{0.5ex}{\fcolorbox[HTML]{FFFFFF}{029E73}{\rule{0pt}{1pt}\rule{5pt}{0pt}}} \SAR ($\epsilon = 3$)
     and 
    \raisebox{0.5ex}{\fcolorbox[HTML]{FFFFFF}{CC78BC}{\rule{0pt}{1pt}\rule{5pt}{0pt}}} \CAR ($\epsilon = 5$)
    \raisebox{0.5ex}{\fcolorbox[HTML]{FFFFFF}{CA9161}{\rule{0pt}{1pt}\rule{5pt}{0pt}}} \SAR ($\epsilon = 5$).
    }
    \label{app:fig:sparsity_real}
\end{figure*}

\textbf{Sparsity}. 
\cref{app:fig:sparsity_synthetic,app:fig:sparsity_real} shows the empirical average sparsity of the causal recourse methods for both synthetic and realistic stochastic processes. 
We do not report values for \IMF since it always acts on \textit{all} the actionable features ($|\calI| = 3$).
In \cref{app:fig:sparsity_synthetic}, \TSAR presents a similar or lower sparsity than other methods. However, as \cref{sec:synthetic-experiment} shows, \TSAR is the only method achieving good validity over time.
Thus, these results suggest that incorporating time might not increase the sparsity of the solutions. 
In the case of approximate SCMs, \cref{app:fig:sparsity_real} shows how \TSAR provides sparse interventions for {\tt Adult}, but increasingly larger interventions for {\tt Loan}.
{\tt COMPAS} has only one actionable feature, thus the sparsity is equal for all approaches. 
As highlighted in \cref{sec:real-experiment}, \TSAR performance is also dependent on the quality of the estimator $\tilde{P}(\vX^t)$.

\begin{figure*}[h]
    \centering
    \begin{minipage}{0.32\textwidth}
        \centering
        \caption*{Linear}
        \includegraphics[width=\linewidth]{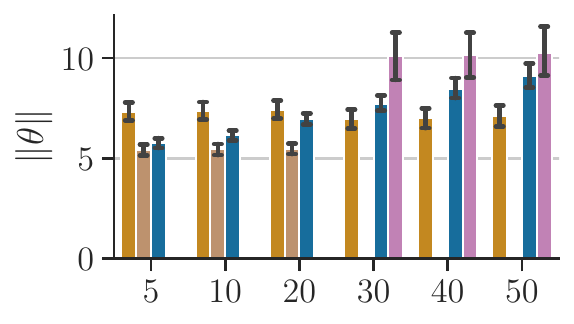}
    \end{minipage}
    \begin{minipage}{0.32\textwidth}
        \centering
        \caption*{Seasonal}
        \includegraphics[width=\linewidth]{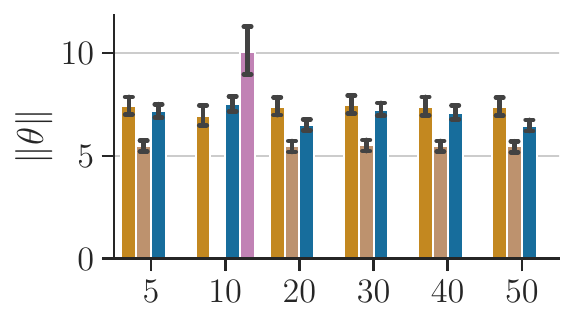}
    \end{minipage}
    \begin{minipage}{0.32\textwidth}
        \centering
        \caption*{Linear+Seasonal}
        \includegraphics[width=\linewidth]{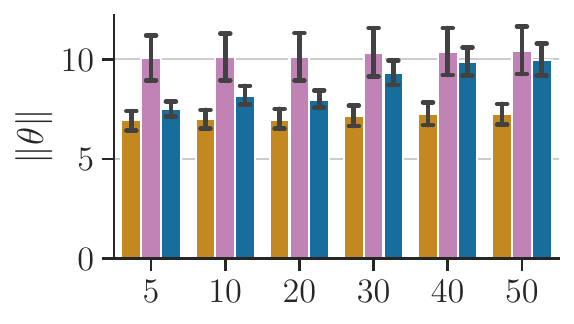}
    \end{minipage}
    \begin{minipage}{0.32\textwidth}
        \centering
        \includegraphics[width=\linewidth]{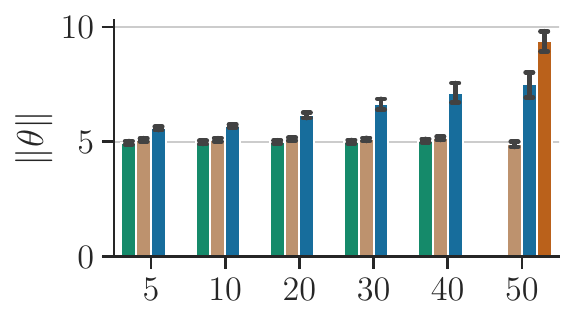}
    \end{minipage}
    \begin{minipage}{0.32\textwidth}
        \centering
        \includegraphics[width=\linewidth]{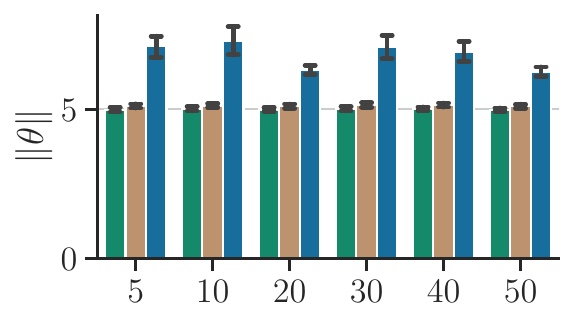}
    \end{minipage}
    \begin{minipage}{0.32\textwidth}
        \centering
        \includegraphics[width=\linewidth]{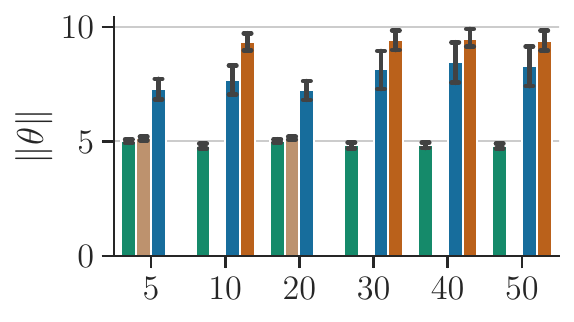}
    \end{minipage}

    \caption{Empirical average cost and standard deviation for the top-3 methods achieving recourse in the Linear (top) and Non-Linear (bottom) ANMs.
    We report the results for all the available trends $m(t) \in \{ \hbox{Linear}, \hbox{Seasonal}, \hbox{Linear+Seasonal} \}$.
    Legend:
    \raisebox{0.5ex}{\fcolorbox[HTML]{FFFFFF}{0173B2}{\rule{0pt}{1pt}\rule{5pt}{0pt}}} \TSAR
    \raisebox{0.5ex}{\fcolorbox[HTML]{FFFFFF}{DE8F05}{\rule{0pt}{1pt}\rule{5pt}{0pt}}} \CAR ($\epsilon = 3$)
    \raisebox{0.5ex}{\fcolorbox[HTML]{FFFFFF}{029E73}{\rule{0pt}{1pt}\rule{5pt}{0pt}}} \SAR ($\epsilon = 3$)
    \raisebox{0.5ex}{\fcolorbox[HTML]{FFFFFF}{D55E00}{\rule{0pt}{1pt}\rule{5pt}{0pt}}} \IMF ($\epsilon = 3$)
     and 
    \raisebox{0.5ex}{\fcolorbox[HTML]{FFFFFF}{CC78BC}{\rule{0pt}{1pt}\rule{5pt}{0pt}}} \CAR ($\epsilon = 5$)
    \raisebox{0.5ex}{\fcolorbox[HTML]{FFFFFF}{CA9161}{\rule{0pt}{1pt}\rule{5pt}{0pt}}} \SAR ($\epsilon = 5$).
    \raisebox{0.5ex}{\fcolorbox[HTML]{FFFFFF}{FBAFE4}{\rule{0pt}{1pt}\rule{5pt}{0pt}}} \IMF ($\epsilon = 5$).
    }
    \label{app:fig:cost_syn}
\end{figure*}

\begin{figure*}[h]
    \centering
    \begin{minipage}{0.32\textwidth}
        \centering
        \caption*{\texttt{Adult}}
        \includegraphics[width=\linewidth]{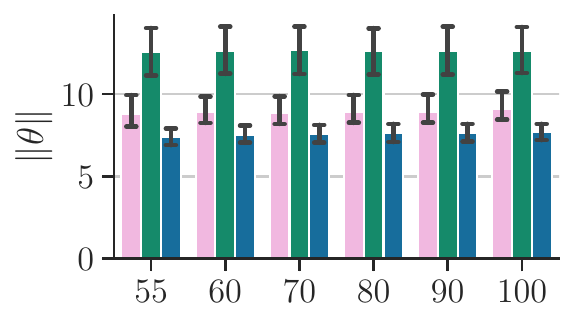}
    \end{minipage}
    \begin{minipage}{0.32\textwidth}
        \centering
        \caption*{\texttt{COMPAS}}
        \includegraphics[width=\linewidth]{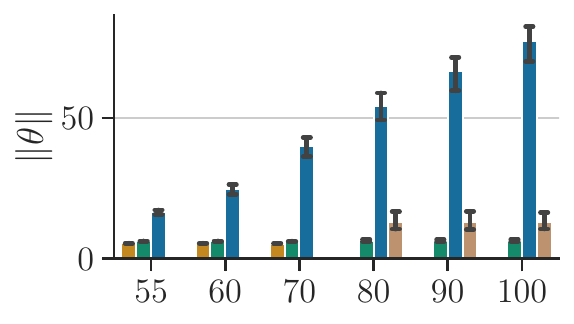}
    \end{minipage}
    \begin{minipage}{0.32\textwidth}
        \centering
        \caption*{\texttt{Loan}}
        \includegraphics[width=\linewidth]{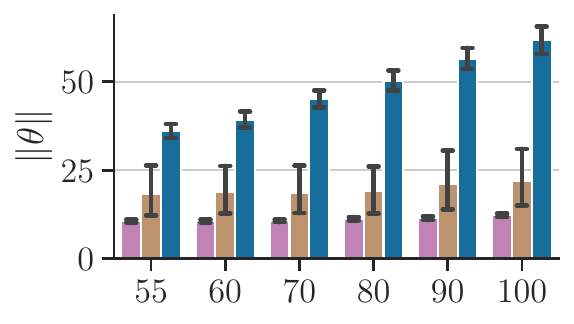}
    \end{minipage}
    
    \caption{Empirical average cost and standard deviation for the top-3 methods achieving recourse in the realistic datasets under a non-linear trend.
    Legend:
    \raisebox{0.5ex}{\fcolorbox[HTML]{FFFFFF}{0173B2}{\rule{0pt}{1pt}\rule{5pt}{0pt}}} \TSAR
    \raisebox{0.5ex}{\fcolorbox[HTML]{FFFFFF}{029E73}{\rule{0pt}{1pt}\rule{5pt}{0pt}}} \SAR ($\epsilon = 0.05$)
     and 
    \raisebox{0.5ex}{\fcolorbox[HTML]{FFFFFF}{CC78BC}{\rule{0pt}{1pt}\rule{5pt}{0pt}}} \CAR ($\epsilon = 0.5$)
    \raisebox{0.5ex}{\fcolorbox[HTML]{FFFFFF}{CA9161}{\rule{0pt}{1pt}\rule{5pt}{0pt}}} \SAR ($\epsilon = 0.5$)
    \raisebox{0.5ex}{\fcolorbox[HTML]{FFFFFF}{FBAFE4}{\rule{0pt}{1pt}\rule{5pt}{0pt}}} \IMF ($\epsilon = 0.5$).}
    \label{app:fig:cost_real}
\end{figure*}

\textbf{Cost.} 
\cref{app:fig:cost_syn,app:fig:cost_real} shows the empirical average cost for the users for which all (non-)causal recourse methods found a valid intervention.
We consider only the top-3 methods achieving recourse for each time step $t$. 
In realistic and synthetic settings, \TSAR can provide \textit{cost-adaptive} interventions which follow the underlying trend.
For example, we can observe this phenomenon in both {\tt COMPAS} and {\tt Loan} (\cref{app:fig:cost_real}).
It is also visible for $m(t) \in \{\hbox{Linear},\hbox{Linear+Seasonal}\}$ in the Non-linear and Linear ANMs, respectively. 
We also notice how \TSAR seems to provide costlier recourses than the standard robust methods.
However, in \texttt{Adult}, \TSAR produces cheaper interventions than the other approaches.
We can explain this behaviour for \texttt{Adult} by looking at the analysis of the successful intervention sets $\calI$ in \cref{sec:real-experiment}. 

In conclusion, by incorporating an estimator of the stochastic process, we can provide \textit{sparse} interventions more resilient to time.
These interventions tend to be costlier and the cost varies with the time lag $\tau$ when they will be applied.
However, robust (non-)causal methods achieve dissimilar validity, thus making them not fully comparable to each other by measuring their cost.
Nevertheless, we believe the analysis has merit since it hints at a tradeoff between sparsity, cost and validity. 

\section{Further Experiments With a Perfect Estimator}
\label{app:experimental-setting-real}

\begin{figure*}[h]
    \centering
    \includegraphics[width=0.9\linewidth]{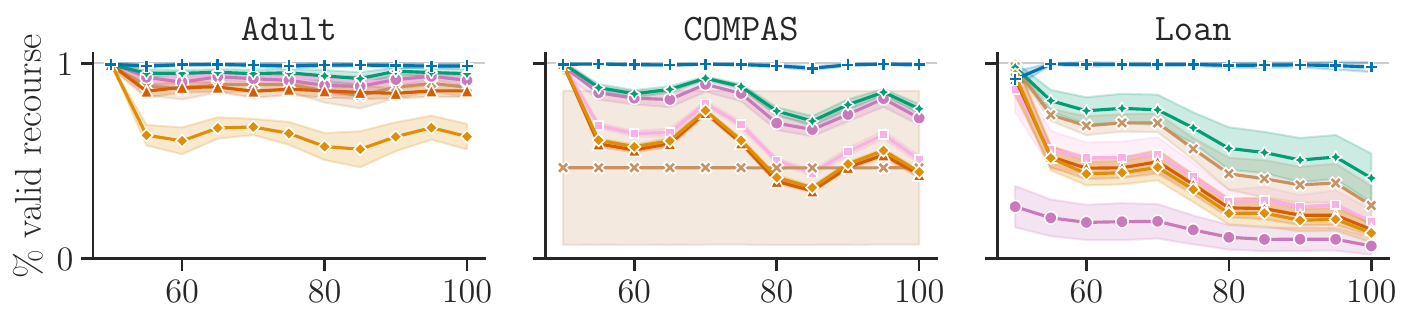}
    \caption{\textbf{Effect of time on realistic datasets.}
Empirical average validity and standard deviation (10 runs) for the robust  ($\epsilon \in \{0.05, 0.5\}$) and time-aware causal recourse methods for the realistic datasets under a non-linear trend.
Legend:
    \raisebox{0.5ex}{\fcolorbox[HTML]{FFFFFF}{0173B2}{\rule{0pt}{1pt}\rule{5pt}{0pt}}} \TSAR
    \raisebox{0.5ex}{\fcolorbox[HTML]{FFFFFF}{DE8F05}{\rule{0pt}{1pt}\rule{5pt}{0pt}}} \CAR ($\epsilon = 0.05$)
    \raisebox{0.5ex}{\fcolorbox[HTML]{FFFFFF}{029E73}{\rule{0pt}{1pt}\rule{5pt}{0pt}}} \SAR ($\epsilon = 0.05$)
    \raisebox{0.5ex}{\fcolorbox[HTML]{FFFFFF}{D55E00}{\rule{0pt}{1pt}\rule{5pt}{0pt}}} \IMF ($\epsilon = 0.05$)
     and 
    \raisebox{0.5ex}{\fcolorbox[HTML]{FFFFFF}{CC78BC}{\rule{0pt}{1pt}\rule{5pt}{0pt}}} \CAR ($\epsilon = 0.5$)
    \raisebox{0.5ex}{\fcolorbox[HTML]{FFFFFF}{CA9161}{\rule{0pt}{1pt}\rule{5pt}{0pt}}} \SAR ($\epsilon = 0.5$)
    \raisebox{0.5ex}{\fcolorbox[HTML]{FFFFFF}{FBAFE4}{\rule{0pt}{1pt}\rule{5pt}{0pt}}} \IMF ($\epsilon = 0.5$).
}
    \label{app:fig:task-3-ground-truth}
\end{figure*}

We replicated the experiments in \cref{sec:real-experiment} by using instead the perfect estimator $\tilde{P}(\vX^t) = P(\vX^t)$ of the stochastic process for each dataset.
\cref{app:fig:task-3-ground-truth} shows how \TSAR offers superior performances in terms of validity than robust (non-)causal algorithmic recourse methods. 
In \texttt{COMPAS} and \texttt{Loan}, \TSAR achieves now perfect validity over all time steps.
These results highlight the importance of relying on a good estimator of the stochastic process and, as outlined in \cref{sec:conclusion-limitations}, we argue it is also a mandatory requirement for realistic applications of the proposed method. 

\section{Additional Theoretical Results on Cost Stability}
\label{app:additional_result_cost_stability}

In this section, we provide an upper bound on the cost stability of recourse suggestions under a more expressive cost function.

Previous research has shown how providing recourse without considering the user's preferences can lead to sub-optimal interventions \citep{detoni2024personalized}.
This is why \textit{personalized} AR, in line with multi-attribute decision making \citep{Keeney1993, pigozzi2016preferences}, models the cost function as an \textit{additive independence model} $C(\hat{\vx}, \vx) = \vw^{\top}|\hat{\vx} -  \vx|$, where the weights $\vw \in \bbR^d$ encapsulate the user's preferences \citep{detoni2023synthesizing,detoni2024personalized}.
We assume we can learn these weights, either from historical data, \eg surveys and interviews \citep{rawal2020beyond}, or by interacting with the end-user \citep{detoni2024personalized}.
In the following, we explicitly consider the evolution of the user's preferences $\vW$, as it also impacts the effectiveness of recourse, although doing so can be avoided for non-personalized AR approaches.

We assume the user preferences can be represented as a stochastic process $P(\vW^t)$.
We do not put any prior assumption on how $P(\vW^t)$ factorizes, since it is not relevant for our results. In line with previous work \citep{detoni2024personalized}, we could imagine $P(\vW^t)$ follows a causal model.
Then, we can provide the following upper bound:

\begin{theorem}
    Consider the discrete-time stochastic processes $P(\vX^t, \allowbreak Y^t \allowbreak)$, $P(\vW^t)$ and a parametrized cost function $C(\hat{\vx}, \vx; \vw) = \dprod{|\hat{\vx}-\vx|}{\vw}$ with bounded $-k \leq w_i^t, X_i^t \leq k$ for $k \in \bbR^+$.
    Given a realization $\vx^t$ and user's preferences $\vw^t$, the variation of the cost of an intervention $\vtheta$ is upper bounded by:
    \begin{equation*}
        \begin{aligned}
        &\bbE\left[\left|C(\hat{\vx}^{t+\tau}, \vx^{t+\tau}; \vw^{t+\tau}) - C(\hat{\vx}^{t}, \vx^t; \vw^{t})\right|\right] \\
        & \qquad \leq k\sqrt{d} \cdot \bbE\left[ \norm{\vw^{t+\tau}-\vw^t} + \norm{|\hat{\vx}^{t+\tau}-\vx^{t+\tau}|-|\hat{\vx}^t-\vx^t|} \right]
        \end{aligned}
    \end{equation*}
    where $\hat{\vx}^t \sim P^{do(\vtheta)}(\vX^{t} \mid \vX_{nd(\calI)}^{t} = \vx_{nd(\calI)}^{t})$ and $\hat{\vx}^{t+\tau} \sim P^{do(\vtheta)}(\vX^{t+\tau} \mid \vX_{nd(\calI)}^{t+\tau} = \vx_{nd(\calI)}^{t+\tau})$.
    \label{theorem:cost-variation-recourse}
\end{theorem}

\cref{theorem:cost-variation-recourse} shows how the recourse cost changes based on how the users' preferences evolve, and also over the relative difference of the proposed changes given the starting value. 

\begin{proof}
We first apply the following substitutions (a) $\vw' = \vw^{t+\tau}$ and $\hat{\vx}' = \hat{\vx}^{t+\tau}$ (b) $\vx' = \vx^{t+\tau}$ $\hat{\vx}'' = \hat{\vx}^{t}$ (c) $\vw = \vw^{t}$ and $\vx'' = \vx^{t}$, to improve the clarity of the proof.
Further, consider $\Delta C = \expect{\abs{C(\vx, \vx; \vw') - C(\vx, \vx; \vw)}}$. 
Then, the proof is the following:
\begin{align*}
         \Delta C &= \expect{\abs{\dprod{\abs{\vx'-\vx}}{\vw'} - \dprod{\abs{\vx-\vx}}{\vw}}} \\
        & = \expect{\abs{
            \dprod{\abs{\hat{\vx}'-\vx'}-\abs{\hat{\vx}''-\vx''}}{\vw'} + \dprod{\abs{\vw'-\vw}}{\abs{\hat{\vx}''-\vx''}}
        }} \\
        & \overset{(i)}{\leq} \expect{
            \norm{\abs{\hat{\vx}'-\vx'}-\abs{\hat{\vx}''-\vx''}} \cdot \norm{\vw'}
        }  \\
        & \qquad \quad + \expect{
            \norm{\abs{\vw'-\vw}} \cdot \norm{\abs{\hat{\vx}''-\vx''}}
        } \\
        & \overset{(ii)}{\leq} k\sqrt{d}\cdot \expect{\norm{\abs{\hat{\vx}'-\vx'}-\abs{\hat{\vx}''-\vx''}}} \\
        & \qquad \quad + k\sqrt{d} \cdot \expect{\norm{\abs{\vw'-\vw}}} \\
        & = k\sqrt{d}\cdot \expect{
        \norm{\abs{\hat{\vx}'-\vx'}-\abs{\hat{\vx}''-\vx''}} + \norm{\abs{\vw'-\vw}}
        } \\
        & = k\sqrt{d} \cdot \bbE\left[ \norm{\vw^{t+\tau}-\vw^t} + \norm{|\hat{\vx}^{t+\tau}-\vx^{t+\tau}|-|\hat{\vx}^t-\vx^t|} \right]
\end{align*}

where (\textit{i}) follows from the Cauchy-Schwarz inequality, and (\textit{ii}) from the bounds we placed on $\vX$ and $\vW$. On the last step, we reorder the terms and we substitute the temporary variables with the original values.
\end{proof}

\end{document}